\pdfoutput=1

\documentclass[11pt]{article}

\usepackage[final]{acl}

\usepackage{times}
\usepackage{latexsym}
\usepackage{graphicx}
\usepackage{booktabs}
\usepackage{amsmath}
\usepackage{comment}

\usepackage[T1]{fontenc}

\usepackage[utf8]{inputenc}

\usepackage{microtype}

\usepackage{inconsolata}

\usepackage{graphicx}

\usepackage{subcaption}

\usepackage{xr}

\title{The Effect of Surprisal on Reading Times in Information Seeking and Repeated Reading}

\author{Keren Gruteke Klein$^1$, Yoav Meiri$^1$, Omer Shubi$^1$, Yevgeni Berzak$^{1,2}$ \\
 $^1$Faculty of Data and Decision Sciences, \\
 Technion - Israel Institute of Technology, Haifa, Israel \\
 $^2$Department of Brain and Cognitive Sciences, \\
 Massachusetts Institute of Technology, Cambridge, USA \\
 \texttt{\{gkeren,meiri.yoav,shubi\}@campus.technion.ac.il},
\texttt{berzak@technion.ac.il} \\
}

\begin{document}
\maketitle
\begin{abstract}

The effect of surprisal on processing difficulty has been a central topic of investigation in psycholinguistics. Here, we use eyetracking data to examine three language processing regimes that are common in daily life but have not been addressed with respect to this question: information seeking, repeated processing, and the combination of the two. Using standard regime-agnostic surprisal estimates
we find that the prediction of surprisal theory regarding the presence of a linear effect of surprisal on processing times, extends to these regimes. However, when using surprisal estimates from regime-specific contexts that match the contexts and tasks given to humans, we find that in information seeking, such estimates do not improve the predictive power of processing times compared to standard surprisals. Further, regime-specific contexts yield near zero surprisal estimates with no predictive power for processing times in repeated reading. These findings point to misalignments of task and memory representations between humans and current language models, and question the extent to which such models can be used for estimating cognitively relevant quantities. We further discuss theoretical challenges posed by these results.\footnote{Code is available at \href{https://github.com/lacclab/surprisal-non-ordinary-reading}{https://github.com/lacclab/surprisal-non-ordinary-reading}.}

\end{abstract}

\section{Introduction}

A key question in psycholinguistics concerns the cognitive processes that underlie the real-time integration of new linguistic material with previously processed linguistic context. A central framework for examining this question is surprisal theory \cite{hale2001probabilistic,levy2008expectation}. This theory ties word processing cost to the word's surprisal, and 
predicts a linear relation between surprisal and processing difficulty.
Due to its theoretical implications (see \citet{shain2024large} for an extended discussion), multiple studies have tested this prediction empirically with different behavioral methodologies (e.g. eyetracking and self paced reading), corpora (among others, Dundee \cite{dundee}, Natural Stories \cite{futrell2021natural}, MECO \cite{meco2022} and CELER \cite{celer2022}), language models, and languages 
\cite{smith2013,goodkind2018,wilcox2020,brothers2021,berzaklevy2023,wilcox2023surp11,shain2024large,Hoover2023,xu-2023-linearity}. All these studies found significant surprisal effects on processing times. With the exception of \citet{Hoover2023} and \citet{xu-2023-linearity} who obtained evidence for superlinear effects, these studies found a linear relation between surprisal and processing times.

However, thus far this relation has been examined only in one reading regime, which can be referred to as \emph{ordinary reading}. This regime presupposes that the comprehender did not have prior, or at least recent, exposure to the linguistic material. It further assumes that they have no specific goals beyond general comprehension of this material. These assumptions do not hold in many daily situations, where language comprehenders often have \emph{specific goals} with respect to the linguistic input, \emph{process the same input multiple times}, or both. This limits the generality of the conclusions that can be drawn from prior studies.  

In this work, we examine the effect of surprisal on reading times in English L1 in three common, but understudied language processing regimes: (1) information seeking, (2) repeated processing, and (3) the combination of the two. Prior work on information seeking \cite{hahn2023,shubi2023eye} and repeated reading \cite{hyona1990repeated,raney1995freq,meiri2024dejavu} has shown substantial differences in eye movement patterns in these regimes compared to ordinary reading, and the extent to which the predictions of surprisal theory hold in these regimes is currently unknown. 

We analyze and compare the functional form and predictive power of two types of contexts, standard regime-agnostic contexts that capture the general predictability of a word, and regime-specific contexts which include the task in information seeking and a prior appearance of the linguistic content in repeated reading. We examine two main hypotheses stemming from surprisal theory. (1) The presence and functional form of surprisal effects for standard surprisal estimates should extend non-ordinary reading regimes. (2) Surprisal estimates from regime-specific contexts should yield higher predictive power for processing times in the respective regimes compared to regime-agnostic contexts, due to a more accurate representation of the context and the processing goals, which should lead to better alignment with subjective word probabilities.

Our main results are the following:
\begin{enumerate}
    \item  \textbf{Regime-agnostic contexts} yield robust linear surprisal effects in information seeking, repeated reading and their combination, albeit with lower predictive power compared to ordinary reading.
    \item \textbf{Regime-specific contexts} that better match the contexts and tasks given to humans, do not improve the predictive power of surprisal for reading times compared to standard regime-agnostic contexts. 
    \begin{enumerate}
        \item In information seeking, providing the information seeking task in the context does not improve model predictive power for reading times.
        \item In repeated processing, providing models with a prior appearance of the linguistic material leads to in-context memorization, with surprisal values that are close to zero and no predictive power for reading times.
    \end{enumerate}
\end{enumerate}

\section{Related Work}
\label{sec:related-work}

The first studies to empirically examine the relation between surprisal and reading times were \citet{smith2008optimal,smith2013}. They used broad coverage eyetracking and self-paced reading data for English, and found evidence for a linear relation. Following this work, several studies obtained similar results using additional corpora, languages and different methodologies for curve fitting and testing linearity, including \citet{goodkind2018}, \citet{wilcox2020}, \citet{shain2024large} and \citet{wilcox2023surp11}. \citet{Hoover2023} and \citet{xu-2023-linearity} obtained evidence for superlinearity. \citet{brothers2021} found a linear relation in word probability using a controlled self-paced reading experiment and cloze estimates of word probabilities. Re-analysis of this data with language model probabilities resulted in a linear relation in surprisal \cite{shain2024hsp}. Our study continues this line of work and extends it to different reading regimes.

Both information seeking and repeated reading have received limited attention in psycholinguistics. Work that examined information seeking \cite{hahn2023,shubi2023eye} found substantial differences in eye movement patterns compared to ordinary reading. The differences were shown to be driven by the division to task-relevant and task-irrelevant information. Different eye movement behavior was also found in repeated reading, where among others, shorter reading times and longer saccades were observed \cite{hyona1990repeated,raney1995freq}. While the presence and magnitude of surprisal effects in information seeking and repeated reading was previously established \cite{shubi2023eye,meiri2024dejavu}, their functional form and predictive power are yet to be determined.

Multiple studies have pointed out divergences between surprisal estimates and human next word expectations \cite{smith2011cloze,jacobs2020human,ettinger2020bert,eisape2020}, as well as an inverse relationship between the quality of recent language models (as measured by perplexity) and their fit to reading times \cite{Oh2022,shain2024large}. Closest to our work is \citet{vaidya2023humans}, who found that in a repeated reading cloze task, language models have substantially higher next word prediction accuracy compared to humans. They further identified ``induction heads'', which are attention heads that recognize repeated token sequences and increase the probability of the previously observed continuation \cite{elhage2021mathematical}, as a core contributor to this behavior in language models. 
Our findings for repeated reading are in line with these results.

\section{Data}

We use OneStop, an extended version of the dataset by \citet{malmaud2020}, with eye movements from 360 English L1 readers, recorded with an Eyelink 1000+ eyetracker (SR Research). The experiment was conducted under an institutional IRB protocol, and all the participants provided written consent before participating in the study. The textual materials are taken from OneStopQA \cite{berzak2020starc} and comprise 30 articles from the Guardian with 4-7 paragraphs (162 paragraphs in total). Each paragraph in OneStopQA is accompanied by three reading comprehension questions. The textual span in the paragraph which contains the essential information for answering the question correctly, called the critical span, is manually annotated in each paragraph for each question.

An experimental trial consists of reading a single paragraph on a page, followed by answering one reading comprehension question on a new page without the ability to go back to the paragraph. \\
\textbf{Ordinary reading vs information seeking} 180 participants are in an ordinary reading regime in which they see the question only after having read the paragraph. 
The remaining 180 participants are in an information-seeking regime in which the question (but not the answers) is presented prior to reading the paragraph. \\
\textbf{First vs repeated reading} Each participant 
reads 10 articles in a random presentation order, followed by two articles that are presented for a second time with identical text but with a different question for each paragraph. The article in position 11 is a repeated presentation of the article in position 10. The article in position 12 is a repeated presentation of one of the articles in positions 1--9. Thus, OneStop contains both consecutive and non-consecutive repeated reading at the article level.\footnote{ Note that for articles 10 and 11, there are 3-6 intervening paragraphs between the two readings of a paragraph.} 

OneStop has 2,532,799 data points (i.e. word tokens over which eyetracking data was collected). We exclude words that were not fixated, words with a total reading time greater than 3,000 ms, words that start or end a paragraph, words with punctuation, and surprisal values greater than 20 bits. After these filtering steps, we remain with 1,157,609 data points: 541,875 in ordinary reading, 474,674 in first reading information seeking, 82,357 in repeated ordinary reading, and 58,703 in repeated reading information seeking. 

\section{Methodology}
\label{sec:methodology}

We examine four different reading regimes that take advantage of the experimental manipulations in OneStop and reflect different types of interactions with the text. The first is ordinary reading during the first presentation of the text. This regime corresponds to the standard experimental setup in reading studies. Additionally, new to this work, we examine information seeking during first reading, and both ordinary reading and information seeking during repeated text presentation.

We estimate the functional form of the relation between surprisal and reading times using Generalized Additive Models \cite[GAMs,][]{Hastie1986}, which can fit non-linear relations between predictors and responses. We predict word reading times from surprisal and two control variables that were shown to be predictive of reading times above and beyond surprisal: word frequency and word length \cite{kliegl2004,clifton2016eye}. To account for spillover effects  \cite{rayner1998eye}, our models also include the surprisal, frequency and length of the previous word. 

Following prior work \cite[e.g.][]{wilcox2023surp11} our primary reading time measure is \textbf{first pass Gaze Duration}; the time from first entering a word to first leaving it during first pass reading. This measure is associated with the processing difficulty of a word given left-only context and is thus especially suitable for benchmarking against surprisal. In the Appendix, we examine additional measures: Gaze Duration and Total Fixation Duration. For completeness, we also provide results for first pass First Fixation duration and First Fixation duration, which tend to have small surprisal effects and are associated with lexical processing \cite{clifton2007eye,berzaklevy2023}. 
Definitions of all the measures are in section  \ref{sec:additional-measures} in the Appendix.

Surprisal, defined as $-\log p(w_i|w_{<i})$, where $w_i$ is the current word and $w_{<i}$ is the preceding context, is estimated using a language model (see Section \ref{subsec:lms}). The language models we use provide a distribution over sub-words (tokens). We therefore sum the sub-word probabilities to obtain the word's probability. Frequency is defined as $-\log p(w_i)$, using word counts from Wordfreq \cite{robyn_speer_2018}. Word length is measured in number of characters.

We define three models of interest:\footnote{All the models were fitted using \texttt{mgcv} (v1.9.1) \texttt{gam} \cite{wood2004stable} function with cubic splines (\texttt{``cr''}). The models do not include random effects due to convergence issues.}
\begin{itemize}
\item \textbf{Baseline model} which predicts reading times of the current word from the control variables frequency and length and their interaction using tensor product terms \texttt{te}.\footnote{Model formula in R: \\
$RT \sim te(freq, len) + te(freq\_prev, len\_prev)$}
\item \textbf{Linear model} which includes the baseline model terms and linear terms for the surprisal of the current and the previous words.\footnote{Model formula in R: $RT \sim surp + surp\_prev+ te(freq, len) + te(freq\_prev, len\_prev)$}
\item \textbf{Non-linear model} which includes the baseline model terms and smooth terms \texttt{s} for the surprisal of the current and previous words.\footnote{Model formula in R:\\  $RT \sim s(surp, k=6) + s(surp\_prev, k=6) \\+ te(freq, len) + te(freq\_prev, len\_prev)$. The value for $k$ is chosen based on prior work \cite{wilcox2023surp11}.}
\end{itemize}

\subsection{Analysis 1: GAM Visualization}

In this analysis, we visualize the relationship between surprisal and reading times using the linear and non-linear models. If the less constrained non-linear fit is visually similar to the linear fit, this would provide initial evidence for a linear relation between surprisal and reading times. To this end, we fit each of the two models on the reading time data of each of the four reading regimes, and predict reading times for surprisal values in the range of 0-20 in 0.1 increments. We note that differently from some of the prior work that used similar methods \cite{smith2013,wilcox2020,wilcox2023surp11}, we do not average reading times across participants before fitting the models.

\subsection{Analysis 2: Predictive Power}

Complementary to analysis 1, we measure the increase in model log-likelihood relative to the baseline model, which includes only the control variables frequency and length, without surprisal, for both the linear and the non-linear models. A statistically significant difference in the predictive power of the non-linear and linear models would provide evidence against linearity. Following prior work \cite[e.g.][]{wilcox2020,Oh2022,wilcox2023surp11}, we measure predictive power for data point $i$ using delta log-likelihood:
\[
\Delta LL_i = \log L^{target}(RT_i | x^{target})
\]
\[
- \log L^{baseline}(RT_i | x^{baseline})
\]
where $RT_i$ is the reading measure of a single participant over a word, $x^{baseline}$ are the control predictors and $x^{target}$ are the target predictors, which include the control predictors and surprisal. $L^M$ is the likelihood under the model M:
\[
L^M(RT_i|x) = f_{norm}(RT_i | \mu=\hat{RT_i}, \sigma^2 =\sigma_{RT}^2)
\]
where $\hat{RT_i}$ is the RT prediction of the model $M$ given the predictor set $x$, $\sigma_{RT}^2$ is the standard deviation of the residuals of the fitted GAM model $M$ and $f_{norm}$ is the Gaussian density function.

We examine $\Delta LL$, the per-word mean of $ \Delta LL_i$. To reduce the risk of overfitting, we measure $\Delta LL$ on held-out data, using 10-fold cross-validation. A positive $\Delta LL$ indicates that the addition of surprisal terms increases the predictive power of the GAM model. We then compare the $\Delta LL$ of the linear and non-linear GAM models. If there is no significant difference between the two, we do not reject the null hypothesis of a linear relation between surprisal and reading times. Following \citet{wilcox2023surp11}, we test the significance of the differences in the $\Delta LL$ of the two models using a paired permutation test.

\begin{figure*}[ht!]
    \centering
    \begin{subfigure}[t]{0.49\textwidth}
        \centering
        \includegraphics[width=\columnwidth]{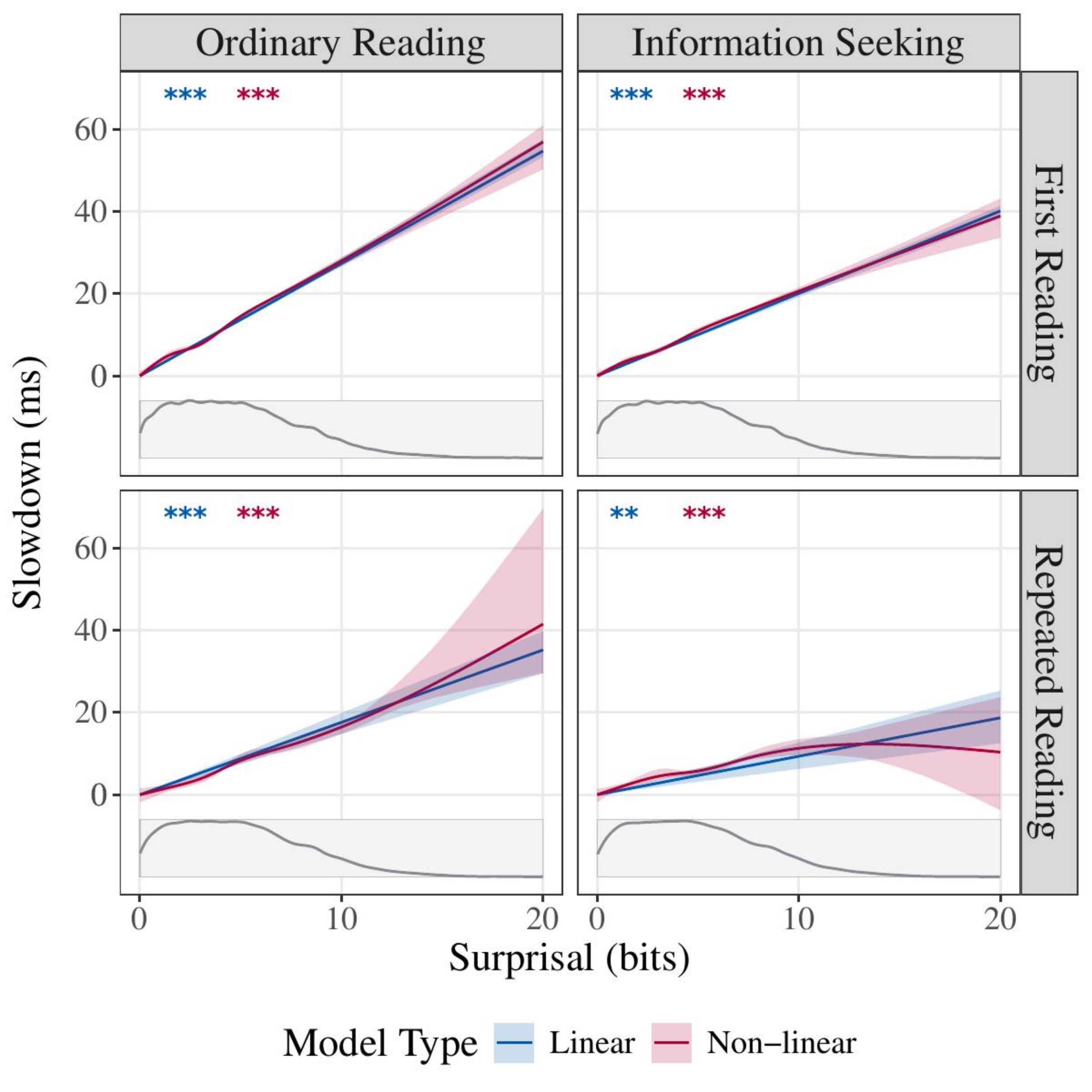}
        \caption{GAM fits for the relation between surprisal and reading times, with bootstrapped 95\% confidence intervals. Top left of each plot, the statistical significance of the \texttt{s} and linear terms of the current word's surprisal. At the bottom of each plot: a density plot of surprisal values.}
        \label{fig:FirstPassGD_GAM}
    \end{subfigure}%
    ~ 
    \begin{subfigure}[t]{0.49\textwidth}
        \centering
        \includegraphics[width=\columnwidth]{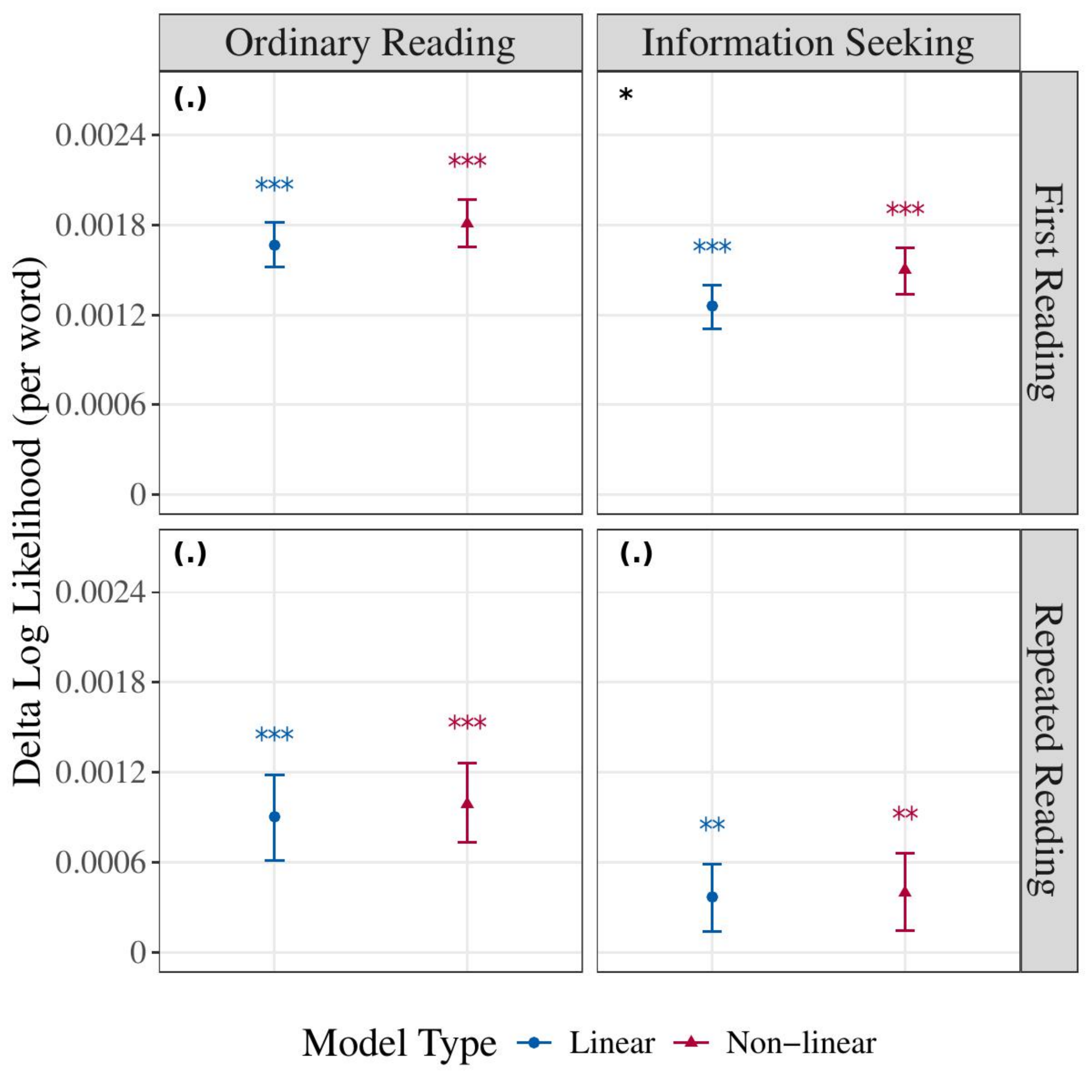}
        \caption{$\Delta LL$ means with 95\% confidence intervals on held-out data using 10-fold cross validation. Above each confidence interval: the statistical significance of a permutation test that checks if the $\Delta LL$ is different from zero. Top left of each plot: the statistical significance of a permutation test for a difference between the $\Delta LL$ of the linear and non-linear models. }
        \label{fig:FirstPassGD_dll}
    \end{subfigure}
    \caption{(a) GAM fits and (b) $\Delta LL$ for first pass Gaze Duration and Pythia-70m surprisals with standard context, using the linear and non-linear models. `***' $p < 0.001$, `**' $p < 0.01$. `*' $p < 0.05$, `(.)' $p \geq 0.05$. \textbf{Key results}: (a) Approximately linear curves for the non-linear models. (b) No statistically significant differences in the $\Delta LL$ of the linear and non-linear models, with the exception of information seeking in first reading. Smaller $\Delta LL$ in information seeking and repeated reading compared to first reading - ordinary reading for both models.}
    \label{fig:FirstPassGD_GAM_dll}
\end{figure*}

\subsection{Language Models and Surprisal Estimation}
\label{subsec:lms}

An important methodological consideration for our study is the choice of the language model. Our selection criteria for the language model is predictive power, as measured by $\Delta LL$. We measure the predictive power of 30 publicly available language models on the OneStop reading time data, and select the model with the highest predictive power across the four reading regimes. 

We examine models from the GPT-2 \cite{radford_language_2019}, GPT-J \cite{gpt-j}, GPT-Neo \cite{gpt-neo}, Pythia \cite{biderman2023pythia}, OPT \cite{zhang2022opt}, Mistral \cite{jiang2023mistral7b}, Gemma \cite{gemmateam2024gemmaopenmodelsbased} and Llama-2 \cite{touvron2023llama} families, ranging from  70 million to 70 billion parameters. We note that this list includes GPT-2-small, which was used in prior work for similar analyses \cite{Oh2022,shain2024large}.
Figure \ref{fig:perplexity_plots} in the Appendix presents model predictive power as a function of the model's log perplexity measured on the 30 articles of OneStopQA. This comparison yields \textbf{Pythia-70m} as the model with the highest predictive power.\footnote{We note that this figure replicates the results of \citet{Oh2022} regarding the relation between perplexity and predictive power for recent language models, and extends them to non-ordinary reading regimes.} Our main analyses therefore use surprisal estimates from this model. To test the robustness of the results to the choice of language model, in the Appendix we present additional analyses with the remaining 29 models.

Recently, \citet{pimentel2024computeprobabilityword} and \citet{oh2024leading} pointed out inaccuracies in the surprisal estimates of models that are based on a beginning-of-word marking tokenizer, such as the Pythia and GPT families. \citet{pimentel2024computeprobabilityword} further propose a modification in the computation of surprisals in such models. While we use the default surprisal values in the results reported below, we have verified that highly similar results are obtained with the estimation method of \citet{pimentel2024computeprobabilityword}.

\subsection{Contexts}

A cardinal manipulation in our study concerns the context  $w_{<i}$ that is provided to the language model for estimating the probability of the current word $w_{i}$.
We examine three approaches for constructing this context. 
\begin{itemize}
    \item \textbf{Standard Context}: In the first, regime-agnostic approach, which we take in Section \ref{sec:analysis-standard}, the context consists of the words preceding the current word in the paragraph.  
    \item \textbf{Regime Context}: In the second, regime-specific approach, in Section \ref{sec:analysis-regime}, the context depends on the reading regime in that it includes the preceding question in information seeking and the paragraph in repeated reading.
    \item \textbf{Prompting + Regime Context}: An additional variant of the Regime Context in Section \ref{sec:analysis-regime} further includes textual prompts that emulate the instructions given to humans.
\end{itemize}

\section{Surprisal from Standard Context}
\label{sec:analysis-standard}

In our first set of analyses, we follow prior work on ordinary first reading, as well as information seeking and repeated reading \cite{shubi2023eye,meiri2024dejavu}, and use standard, reading regime-agnostic surprisal estimates, which are obtained by conditioning the model on the prior textual material in the paragraph.

\subsection{GAM Visualization}
\label{subsec:gams}

Figure \ref{fig:FirstPassGD_GAM} presents the GAM surprisal curves for the linear and non-linear models. Visual inspection suggests that the non-linear model approximately tracks the linear fit. We further note that consistently with the findings of \citet{shubi2023eye} and \citet{meiri2024dejavu}, surprisal effects, which can be inferred from the slope of the curves, are smaller in information seeking compared to ordinary reading, and smaller in repeated reading compared to first reading. 

Figure \ref{fig:Different Surprisal Estimates grid GAM 1} in the Appendix suggests that the results largely hold across different language models, although some of the models with the lowest perplexity also yield sublinear fits. Figure \ref{fig:pythia-70m_grid_GAM} in the Appendix examines additional reading measures for Pythia-70m, with linear fits for Gaze Duration and Total Fixation duration, and mixed results for first pass First Fixation and First Fixation where we observe sublinear curves in first reading. Overall, most curves of the non-linear models appear to approximate their linear counterparts.

In information seeking, \citet{shubi2023eye} have shown different eye movement patterns within and outside task critical information (the critical span). In repeated reading, \citet{meiri2024dejavu} also showed differences between eye movements in consecutive (article 11) and non-consecutive (article 12) repeated article presentation. Figure \ref{fig:pythia70m-Hunting Reread splits} in the Appendix shows that linearity for first pass Gaze Duration holds both within and outside the critical span in information seeking, and also both with and without intervening articles during repeated reading.

\begin{table*}[h!]
\resizebox{\textwidth}{!}{%
\begin{tabular}[t]{|l|l|ll|ll|}
\hline
\textbf{Regime}                                                                & \textbf{\begin{tabular}[c]{@{}l@{}}Standard \\ Context\end{tabular}} & \textbf{\begin{tabular}[c]{@{}l@{}}Regime \\ Context\end{tabular}} & \textbf{Description}                                                                                                                              & \textbf{\begin{tabular}[c]{@{}l@{}}Prompting + \\ Regime Context\end{tabular}} & \textbf{Prompt Text}                                                                                          \\ \hline
\begin{tabular}[t]{@{}l@{}}First reading \\ Ordinary reading\end{tabular}      & P                                                                    & P              & The preceding words in the paragraph.                                                                                                                                                               & Prompt1 + P                    & Prompt1: "You will now read a paragraph."                                                                                                                                                                                                                                                                                        \\ \hline
\begin{tabular}[t]{@{}l@{}}First reading\\ Information seeking\end{tabular}    & P                                                                    & Q + P          & \begin{tabular}[t]{@{}l@{}}The question followed by the \\ preceding words in the paragraph.\end{tabular}                                                                                          & Prompt1 + Q + P                   & \begin{tabular}[t]{@{}l@{}}Prompt1: "You will now be given a question \\ about a paragraph followed by the paragraph. \\ You will need to answer the question."\end{tabular}                                                                                                                                                    \\ \hline
\begin{tabular}[t]{@{}l@{}}Repeated reading\\ Ordinary reading\end{tabular}    & P                                                                    & P + P          & \begin{tabular}[t]{@{}l@{}}The entire paragraph followed by the \\ preceding words in the same paragraph.\end{tabular}                                                                             & \begin{tabular}[t]{@{}l@{}}Prompt1 + P +  \\ Prompt2 + P \end{tabular}           & \begin{tabular}[t]{@{}l@{}}Prompt1: "You will now read a paragraph."\\ Prompt2: "You will now read the same paragraph\\ again."\end{tabular}                                                                                                                                                                                        \\ \hline
\begin{tabular}[t]{@{}l@{}}Repeated reading\\ Information seeking\end{tabular} & P                                                                    & \begin{tabular}[t]{@{}l@{}}Q' + P +  \\ Q + P \end{tabular} & \begin{tabular}[t]{@{}l@{}}The question for the first reading, \\ followed by the paragraph, the question \\ for the second reading and the preceding \\ words in the same paragraph.\end{tabular} & \begin{tabular}[t]{@{}l@{}}Prompt1 + Q' + P +  \\ Prompt2 + Q + P \end{tabular}   & \begin{tabular}[t]{@{}l@{}}Prompt1: "You will now be given a question \\ about a paragraph followed by the paragraph. \\ You will need to answer the question."\\ Prompt2: "You will now read the same paragraph \\ again with a different question before the \\ paragraph. You will need to answer the question.''\end{tabular} \\ \hline
\end{tabular}%
}
\caption{Standard and regime-specific contexts provided to language models. Q and Q' for two different questions, and P for paragraph. The prompts are similar to those presented to human participants in the reading experiment.}
\label{table:regimes}
\end{table*}

\subsection{Predictive Power}

While visual inspection provides initial evidence for the linearity of reading times in surprisal across reading regimes, we further test this hypothesis by comparing the predictive power of the non-linear model relative to that of the linear model. Figure \ref{fig:FirstPassGD_dll} presents the $\Delta LL$ of the linear and non-linear models for first pass Gaze Duration across the four reading regimes. We find that in three of the four regimes, there is no significant difference between the $\Delta LL$ of the two models. In information seeking - first reading, the difference is significant at $p<0.05$. These results largely support our conclusion from the visual inspection of the GAM curves, that the surprisal - reading times relation is linear in all four regimes. 
We further note, that in line with the effect sizes, the predictive power of standard surprisal estimates is smaller in information seeking compared to ordinary reading, and smaller in repeated reading compared to first reading ($p<0.05$ in all cases using a paired permutation test).

Figure \ref{fig:Different Surprisal Estimates grid dll 1} in the Appendix presents the results for first pass Gaze duration across different language models, suggesting that they are robust to the language model choice. Figure \ref{fig:pythia-70m_grid_dll} in the Appendix presents additional reading measures and further shows that the results mostly extend to Gaze Duration and Total Fixation Duration, while mixed results are obtained for First Fixation measures, with larger $\Delta LL$ for the non-linear model in ordinary reading and information seeking during first reading. Figure \ref{fig:pythia70m-Hunting Reread splits} shows that the linearity of first pass Gaze Duration in surprisal holds both within and outside the critical span in information seeking, as well as for consecutive and non-consecutive article repeated reading. 
Overall, our analysis of $\Delta LL$ favors a linear relation between surprisal and reading times across all four reading regimes.

\section{Surprisal from Regime-Specific Context}
\label{sec:analysis-regime}

\begin{figure*}[ht!]
    \centering
    \begin{subfigure}[t]{0.49\textwidth}
        \centering
        \includegraphics[width = 0.99\columnwidth]{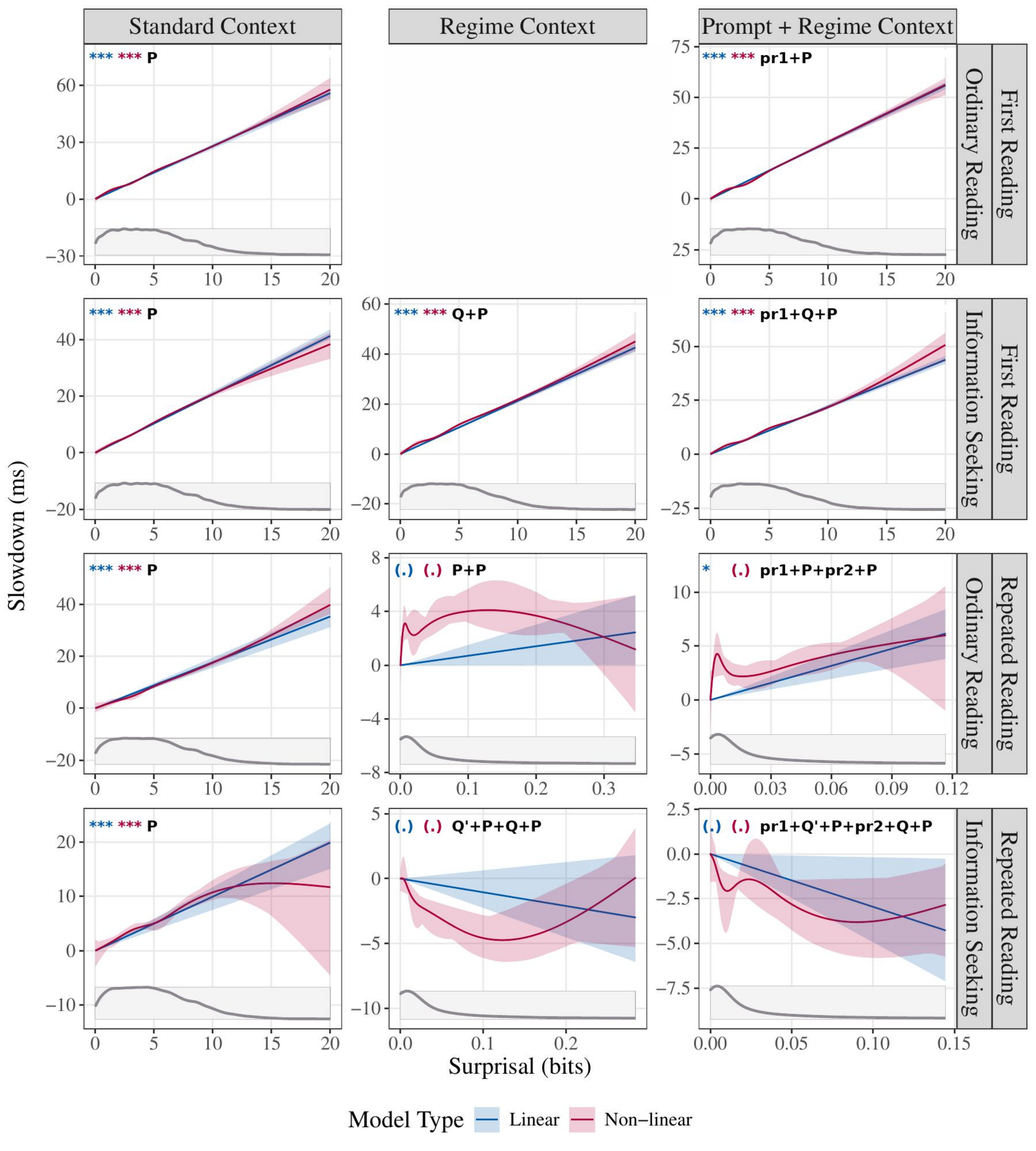}
        \caption{GAM fits for the relation between surprisal and reading times across context types. Slowdown effects in \textit{ms} for first pass Gaze Duration as a function of surprisal, with bootstrapped 95\% confidence intervals. Top left of each plot, the significance of the \texttt{s} and linear terms of the current word's surprisal. At the bottom of each plot: a density plot of surprisal values. 
        \textbf{Key results} for the Regime Context and Prompt + Regime Context: (a) in first reading - information seeking, approximately linear curves for the non-linear model. (b) In the two repeated reading conditions, surprisals are close to zero with no surprisal effect.}
    \label{fig:FirstPassGD_GAM_task_context}
    \end{subfigure}%
    ~ 
    \begin{subfigure}[t]{0.49\textwidth}
        \centering
        \includegraphics[width = 0.99\columnwidth]{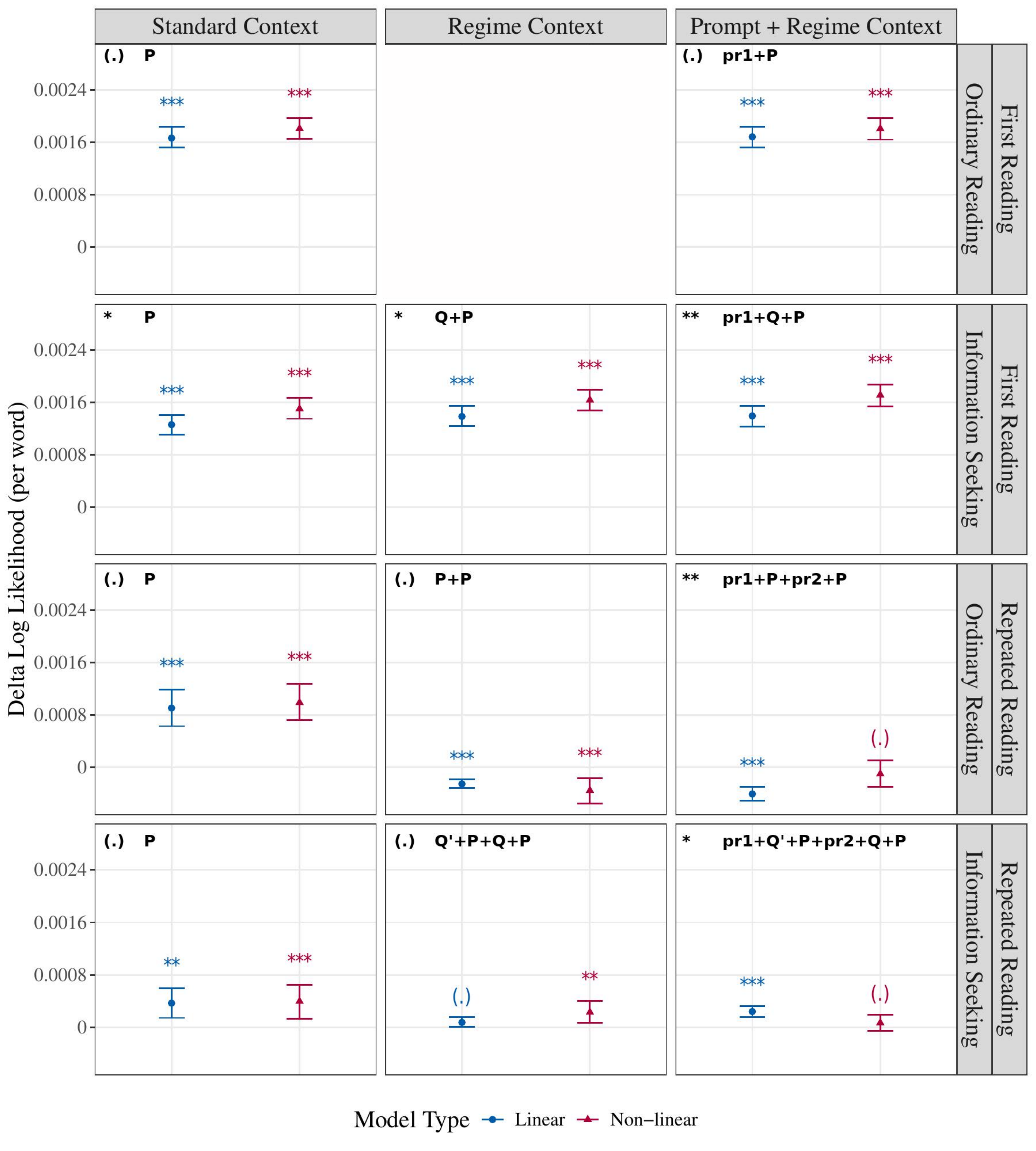}
        \caption{$\Delta LL$ means with 95\% confidence intervals on held-out data using 10-fold cross validation. Above each confidence interval: the statistical significance of a permutation test that checks if the $\Delta LL$ is different from zero. Top left of each plot: significance of a permutation test for a difference between the $\Delta LL$ of the linear and non-linear models.  \textbf{Key results} for Regime Context and Prompt + Regime Context: (1) In first reading - information seeking, no significant differences in the $\Delta LL$ of the linear and non-linear models, and no increase in $\Delta LL$s compared to the Standard Context. (2) In both repeated reading regimes, $\Delta LL$s are \emph{lower} compared to the Standard Context and in most cases not significantly above zero.}
        \label{fig:FirstPassGD_dll_context}
    \end{subfigure}
    \caption{Comparison of GAM fits and $\Delta LL$ for first pass Gaze Duration with surprisal estimates of Pythia-70m from different context types. `***' $p < 0.001$, `**' $p < 0.01$. `*' $p < 0.05$, `(.)' $p \geq 0.05$. }
\label{fig:FirstPassGD_GAM_and_dll_task_context}
\end{figure*}

Thus far, we used surprisal estimates based on the textual context in the paragraph. However, this context does not fully capture the reading task conditioning in the human data. Human participants in the first reading -- information seeking regime receive a question prior to reading the paragraph. In repeated ordinary reading they have already read that paragraph. In repeated reading during information seeking they have previously read the paragraph and received a question prior to both the first and the second reading of the paragraph. These manipulations can alter linguistic expectations and were previously shown to influence reading times  \cite{hyona1990repeated,malmaud2020,shubi2023eye,meiri2024dejavu}. Furthermore, human participants receive explicit instructions regarding the different trial components in the reading experiment.

In the remainder of this work, we compare our results using standard surprisal estimates to surprisal estimates based on context types that more closely match the textual contexts and instructions presented to humans in each of the reading regimes. Our analyses focus on the following questions regarding the three regimes that are not ordinary first reading. (1) Do the linear surprisal effects persist under regime-conditioned surprisal estimates? (2) Do regime-conditioned surprisals lead to better predictive power for human reading times?

To address these questions, in addition to the standard context used in Section \ref{sec:analysis-standard}, we examine three \textbf{regime-specific contexts} that correspond to each of the three reading regimes that involve information seeking and repeated reading. To further enhance the similarity to the experimental setup in the human data, we also examine a variant of the regime contexts in which the model additionally receives \textbf{prompts} that emulate the reading instructions received by human participants. The prompts convey the same content provided in the instructions to human participants in the eyetracking experiment, but are not a verbatim copy, as the original instructions further contain details relevant only for the eyetracking experiment, such as the text triggering targets and button presses associated with each part of the trial. The regime-specific contexts and prompts are presented in Table \ref{table:regimes}. 

We note that although these contexts include the essential components of each reading regime, they do not fully match the eyetracking experiment as they do not include intervening textual material between first and second presentations of a paragraph. This is because the context window of our models is too small to include the text of a full experimental session. To partially address this limitation, in Table \ref{table:article-contexts} in the Appendix we present a prompting scheme for article-level analysis for articles 10 and 11. 
We use this scheme with the 
Pythia-70m model, for which we employ a sliding window mechanism with an overlap size that ensures that each paragraph's first appearance is fully included in the context window of its repeated appearance.

\subsection{GAM Visualization}

In figure \ref{fig:FirstPassGD_GAM_task_context} we present GAM visualizations for the linear and non-linear models. 
We compare surprisals from conditioning on the standard paragraph context P to surprisals from reading regime contexts: Q+P for first reading - information seeking, P+P for repeated reading - ordinary reading, and Q'+P+Q+P for repeated reading - information seeking. We further present results for regime contexts with prompting.

For first reading - information seeking, surprisals from both regime-specific contexts yield linear curves. However, a very different outcome is observed in the repeated reading regimes. In these regimes, there is a collapse of the surprisals to values that are close to zero and 
null effects of surprisal on reading times. Thus, we obtain two different behaviors for information seeking and repeated reading. While the addition of the information seeking task does not substantially alter the predictive power of the model, conditioning twice on the paragraph leads to surprisals that no longer maintain a significant relation to reading times.

\subsection{Predictive Power}

In figure \ref{fig:FirstPassGD_dll_context} we compare the $\Delta LL$ of the linear and non-linear models across standard and regime-specific surprisals with and without prompting. 
In first reading - information seeking, 
the regime context and the prompt + regime context provide weak evidence against linearity ($p=0.04$ and $p=0.01$ respectively).
Crucially, regime conditioning and prompting do not improve predictive power in this regime;
the $\Delta LL$ of the regime context is not significantly higher compared to the standard context ($p=0.25$ linear; $p=0.27$ non-linear, using a paired permutation test). Adding prompting yields similar outcomes compared to the standard context ($p=0.22$ linear; $p=0.08$ non-linear).

In the repeated reading regimes we observe a different pattern. Importantly, the regime contexts in the ordinary reading condition lead to a \emph{decrease} in the $\Delta LL$ compared to the standard context in both the linear ($p=0.001$) and non-linear cases ($p=0.009$). A similar pattern is observed when adding prompting, with $p=0.001$ for the linear model and $p=0.038$ for the non-linear model. 
The regime contexts in the information seeking condition exhibit the same pattern of $\Delta LL$ decrease compared to the standard context, which is significant both without prompting ($p=0.017$ linear; $p=0.004$ non-linear) and with prompting ($p=0.091$ linear; $p=0.027$ non-linear). 
Furthermore, in nearly all cases the regime context $\Delta LL$ is not significantly above zero, suggesting that the corresponding surprisal estimates have no predictive power with respect to reading times. Taken together with the GAM visualizations in Figure \ref{fig:FirstPassGD_GAM_task_context}, we conclude that the examined language models are misaligned with human reading patterns in repeated reading, and do not provide useful surprisal estimates when conditioned for repeated reading. 

These results are consistent across all the models examined, and specifically for the larger models, which could a-priori be expected to be more sensitive to context conditioning and prompting. In the Appendix, we present these results for GPT-2-small in Figure \ref{fig:GPT2-small context} and for the largest Llama and Mistral models, Llama 70b in Figure \ref{fig:Llama 70b context} and Mistral Instruct v0.3 7b in Figure \ref{fig:Mistral Instruct v0.3 7B context}.
Furthermore, Figure \ref{fig:Pythia70m Article context} in the Appendix suggests that they generalize to repeated reading with  intervening paragraphs between the two paragraph presentations for articles 10 and 11. 

\section{Discussion and Conclusion}

Surprisal theory predicts a linear relationship between surprisal and word processing times. This prediction found support in studies with ordinary reading, but was not previously examined in information seeking and repeated reading. We find evidence that with standard surprisal estimates, the prediction of surprisal theory for a linear effect of surprisal on reading times holds in these regimes. We further find that the effect size and predictive power of standard surprisal estimates diminish in information seeking and repeated reading.

Our attempt to improve language model predictive power with regime-specific contexts yields two primary findings. First, we observe that regime-specific surprisal estimates in first reading - information seeking {do not improve} the fit to human reading times. A more severe case of estimation collapse is observed in repeated reading, where we find near zero surprisal estimates with no predictive power for reading times, likely due to in-context memorization. 

These findings highlight two different types of misalignment between language models and humans. Information seeking demonstrates a misalignment in the representation of task information. Repeated reading suggests very different memory and retrieval abilities in humans and current language models. These misalignments question not only the suitability of current language models as cognitive models of human language processing, but also the psycholinguistic relevance of quantities extracted from such models. 

We entertain two possible explanations for the discrepancies in the real-time processing and memory mechanisms of humans and language models. The first explanation is that this mismatch stems from architectural and/or training aspects of current language models. If this is indeed the case, they can be potentially alleviated or even completely resolved with architectural or training procedure changes to said models; it is well possible that future architectures will better capture task relevant information, or handle repeated text in ways that are more commensurate with human processing. 

The second explanation poses a challenge to language processing theory, and in particular to the view of surprisal as a ``causal bottleneck'' for observed behavior \cite{levy2008expectation}. According to this view, whatever the underlying linguistic processing mechanisms and representations may be, their effect on processing times is mediated through surprisal. Although better representation of the context should yield better estimates of subjective surprisals and thus better reflect processing times, we do not observe this in practice. 

One could alternatively argue that factors that come into play in non-ordinary processing regimes and affect reading times either cannot or should not be encoded in surprisals. Surprisal theory accounts only for processing difficulty, while reading times may reflect additional factors of cognitive state, which do not directly speak to processing difficulty (e.g. one may skim through portions of the text because they are less relevant for the comprehender goals, not because they are easier to process). Future empirical and theoretical work is required to make further progress on these questions.

\section{Limitations}
\label{sec:limitations}

Our work has multiple limitations. Due to the lack of eyetracking data for information seeking and repeated reading in other languages, we address only English. The readers are adult native speakers in the age range of 18--52. Additional data collection in other languages, ages and participant groups are needed to establish the generality of the conclusions. 
The experimental design is further constrained to one variant of each reading regime, leaving many other variants unaddressed. For example, an experimental trial consists of a single paragraph. In daily interactions with text, information seeking can be over both shorter and longer textual units. In repeated reading, consecutive reading is at the article level with intervening paragraphs, and doesn't cover immediate repeated reading which involves working memory. In non-consecutive reading, we have at most 10 intervening articles. In both cases, repeated reading can occur more than once. 

Further limitations concern the language models used. The context window of the models available with our computing resources is not sufficient to address non-consecutive article repeated reading, which requires storing up to 12 articles at once in the context provided to the model. Additional work with large context windows is required to fully address the repeated reading experimental design in the eyetracking data.

We use the term ordinary reading to refer to a first reading for comprehension. However, following \citet{huettig2023} we acknowledge that this term is not without faults. Relatedly, while reading comprehension questions are essential for encouraging attentive reading, their presence after each paragraph may lower the ecological validity of the data, especially in the ordinary reading regime. Reading in a lab setting may further limit the applicability of the results to daily reading situations.

\bibliography{custom}

\appendix
\renewcommand{\thefigure}{A\arabic{figure}}
\renewcommand{\thetable}{A\arabic{table}}

\onecolumn

\section{Additional reading time measures}
\label{sec:additional-measures}

\begin{itemize}
    \item First Fixation Duration (FF): the time elapsed from the beginning of the first fixation on a word to the beginning of the next saccade.
    \item First pass First Fixation Duration (first pass FF): the duration of the first fixation on a word during first pass reading. All words that were skipped in the initial pass are ignored when using this measure.
    \item Gaze Duration (GD): the time elapsed from the beginning of the first fixation on a word to the beginning of the first saccade leading to a different word.
    \item Total Fixation duration (TF): the cumulative duration of all fixations on a word (excluding saccades).
\end{itemize}

\section{Prompts for repeated reading with consecutive article presentation (articles 10 and 11)}

\begin{table*}[h!]
\resizebox{\textwidth}{!}{%
\begin{tabular}{llllll}
\hline
\textbf{Regime}                                                                & \textbf{\begin{tabular}[t]{@{}l@{}}Standard \\ Context\end{tabular}}                                                                    & \multicolumn{2}{l}{\textbf{\begin{tabular}[t]{@{}l@{}}Regime \\ Context\end{tabular}}}                                                                                                                                        & \multicolumn{2}{l}{\textbf{\begin{tabular}[t]{@{}l@{}}Prompting + \\ Regime Context\end{tabular}}}                                                                                                                                                                                                                                                                                                              \\ \hline
\begin{tabular}[t]{@{}l@{}}First reading \\ Ordinary reading\end{tabular}      & \begin{tabular}[t]{@{}l@{}}All the preceding paragraphs in the article\\ and the preceding words in the current paragraph.\end{tabular} & \multicolumn{2}{l}{\begin{tabular}[t]{@{}l@{}}All the preceding paragraphs in the article and \\ the preceding words in the current paragraph.\end{tabular}}                                                                  & \multicolumn{2}{l}{\begin{tabular}[t]{@{}l@{}}Prompt: You will now read an article, \\ paragraph by paragraph.\end{tabular}}                                                                                                                                                                                                                                                                                    \\ \hline
\begin{tabular}[t]{@{}l@{}}First reading\\ Information seeking\end{tabular}    & \begin{tabular}[t]{@{}l@{}}All the preceding paragraphs in the article\\ and the preceding words in the current paragraph.\end{tabular} & \multicolumn{2}{l}{\begin{tabular}[t]{@{}l@{}}All the preceding paragraphs and questions in the \\ article and the preceding words in the current paragraph.\end{tabular}}                                                    & \multicolumn{2}{l}{\begin{tabular}[t]{@{}l@{}}Prompt: You will now read an article, \\ paragraph by paragraph.\\ Before each paragraph, you will be given \\ a question that you will need to answer.\end{tabular}}                                                                                                                                                                                             \\ \hline
\begin{tabular}[t]{@{}l@{}}Repeated reading\\ Ordinary reading\end{tabular}    & \begin{tabular}[t]{@{}l@{}}All the preceding paragraphs in the article\\ and the preceding words in the current paragraph.\end{tabular} & \multicolumn{2}{l}{\begin{tabular}[t]{@{}l@{}}The entire article, followed by all the preceding \\ paragraphs in the article and the preceding words in \\ the current paragraph.\end{tabular}}                               & \multicolumn{2}{l}{\begin{tabular}[t]{@{}l@{}}Prompt 1: You will now read an article, \\ paragraph by paragraph.\\ Prompt 2 (between first and second reading): \\ You will now read the same article again.\end{tabular}}                                                                                                                                                                                      \\ \hline
\begin{tabular}[t]{@{}l@{}}Repeated reading\\ Information seeking\end{tabular} & \begin{tabular}[t]{@{}l@{}}All the preceding paragraphs in the article\\ and the preceding words in the current paragraph.\end{tabular} & \multicolumn{2}{l}{\begin{tabular}[t]{@{}l@{}}The entire article with questions, followed by all the \\ preceding paragraphs in the article with questions and \\ the preceding words in the current paragraph.\end{tabular}} & \multicolumn{2}{l}{\begin{tabular}[t]{@{}l@{}}Prompt 1: You will now read an article, \\ paragraph by paragraph. \\ Before each paragraph, you will be given \\ a question that you will need to answer.\\ Prompt 2 (between first and second reading): \\ You will now read the same article again \\ with a different question before each paragraph. \\ You will need to answer the questions.\end{tabular}} \\ \hline
\end{tabular}%
}
\caption{Context types and prompts for consecutive reading of an article in positions 10 and 11, with 3-6 intervening paragraphs between two readings of the same paragraph.}
\label{table:article-contexts}
\end{table*}

\begin{figure*}[ht!]
    \centering
    \begin{subfigure}[t]{0.99\textwidth}
        \centering
        \includegraphics[width=0.99\linewidth]{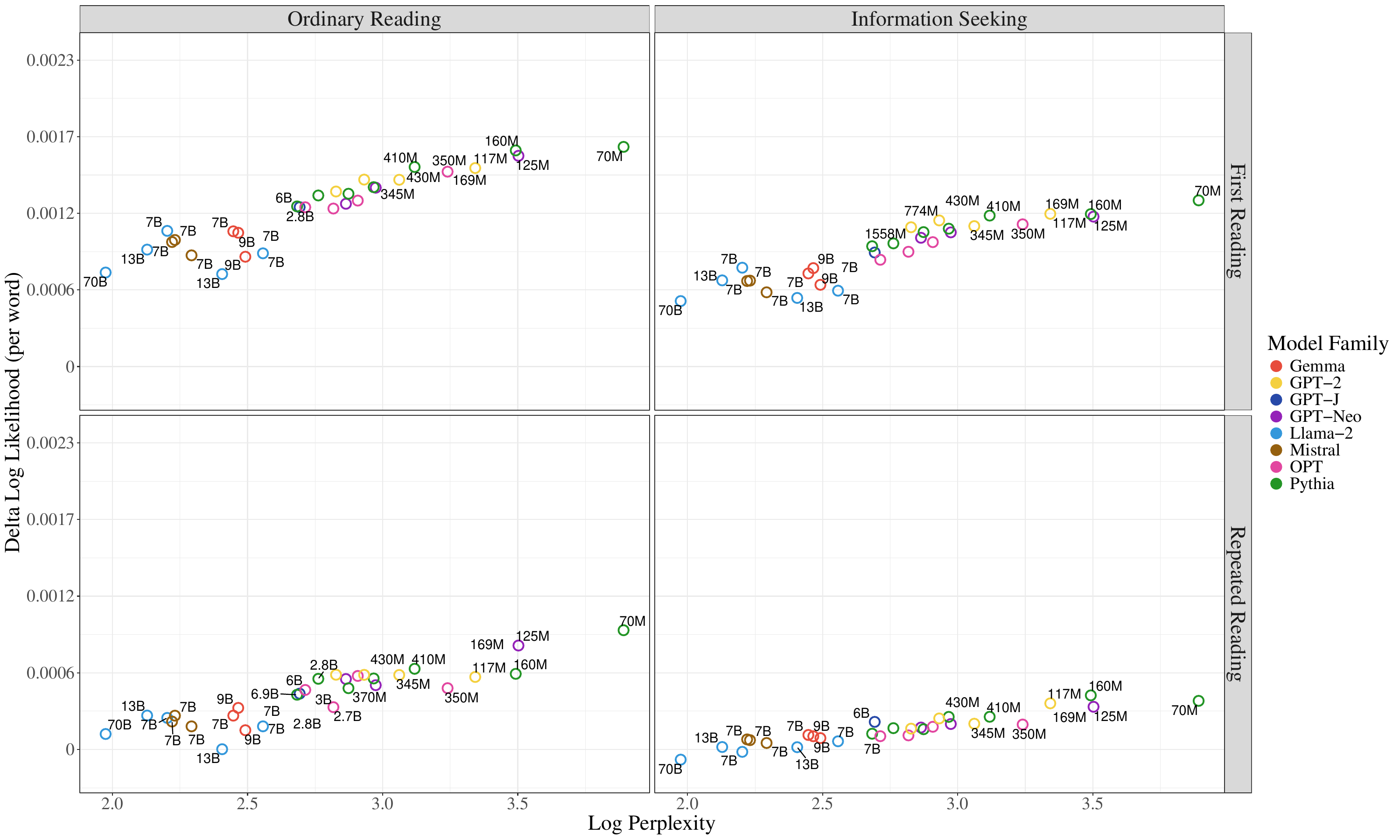}
        \caption{$\Delta LL$ by perplexity for first pass Gaze Duration. $\Delta LL$ of the \emph{linear model} calculated on held-out data using 10-fold cross validation.}
        \label{fig:perplexity plot linear models}
    \end{subfigure}%
    ~ \\
    \begin{subfigure}[t]{0.99\textwidth}
        \centering
        \includegraphics[width=0.99\linewidth]{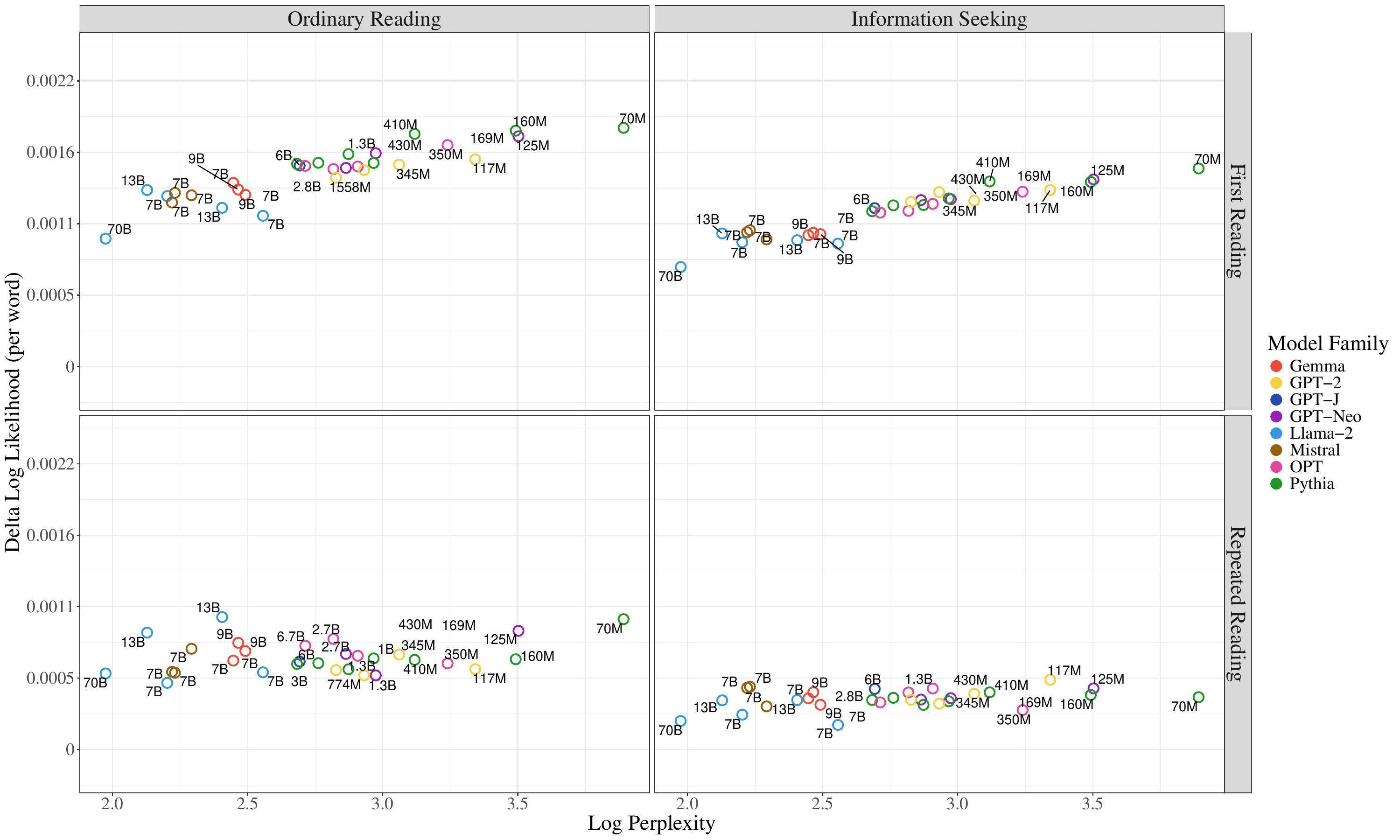}
        \caption{$\Delta LL$ by perplexity for first pass Gaze Duration. $\Delta LL$ of the \emph{non-linear model} calculated on held-out data using 10-fold cross validation.}
        \label{fig:perplexity plot non-linear models}
    \end{subfigure}
    \caption{Predictive power for reading times across different language models as a function of log-perplexity. Perplexity here is sentence-level perplexity averaged over all sentences in OneStopQA (the 30 articles used for the eye-tracking experiment).}
    \label{fig:perplexity_plots}
\end{figure*}

\begin{figure*}[ht!]
    \centering
    \begin{subfigure}[t]{\textwidth}
        \centering
        \includegraphics[width=1.00\linewidth]{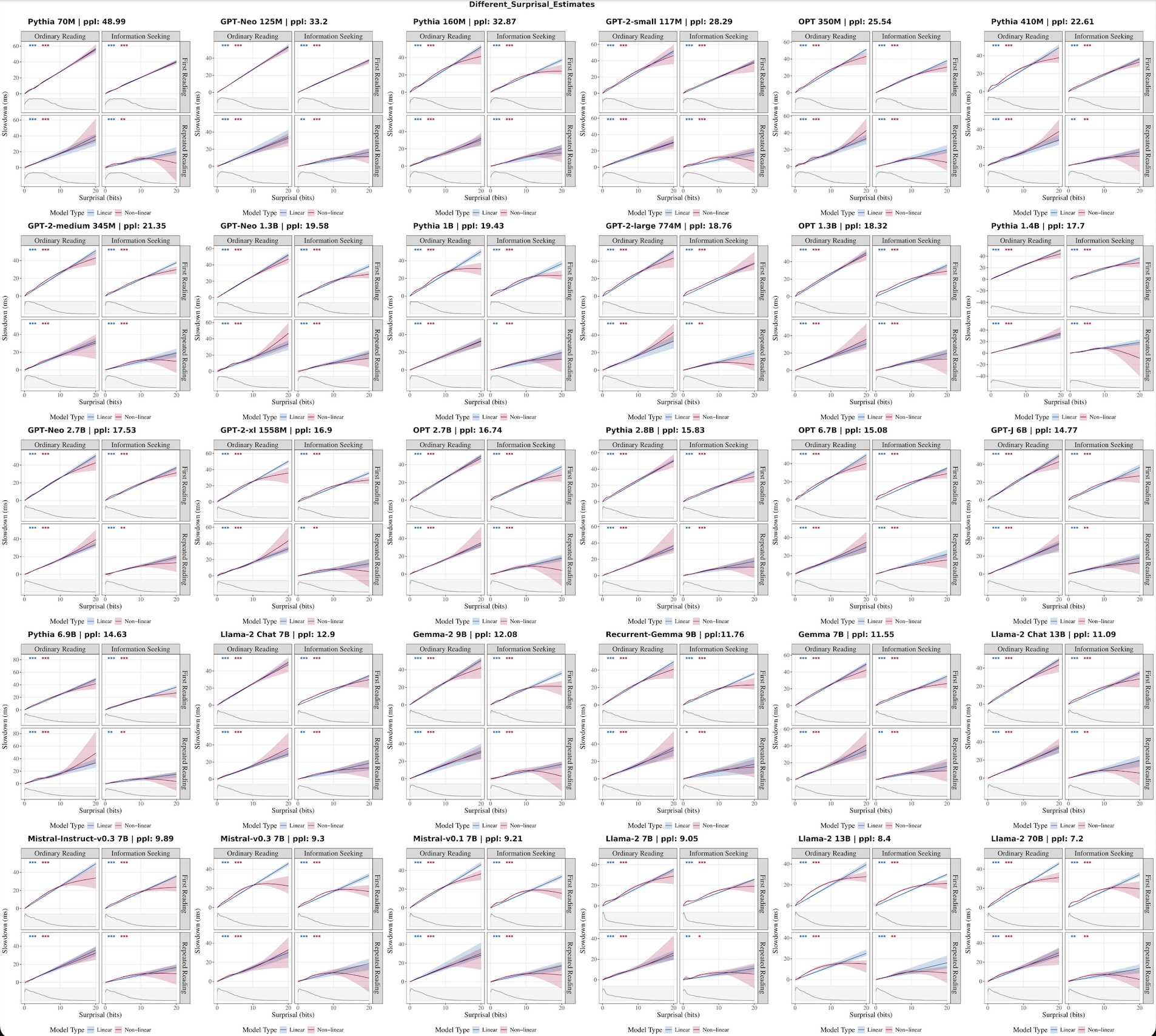}
        \caption{\textbf{GAM fits for the relation between surprisal and reading times}, with bootstrapped 95\% confidence intervals. At the top left of each plot: the significance of the \texttt{s} and linear terms of the current word's surprisal. At the bottom of each plot: a density plot of surprisal values.}
        \label{fig:Different Surprisal Estimates grid GAM 1}
    \end{subfigure}%
\end{figure*}

\begin{figure*}[ht!]\ContinuedFloat
    \centering
    \begin{subfigure}[t]{\textwidth}
        \centering
        \includegraphics[width=1.00\linewidth]{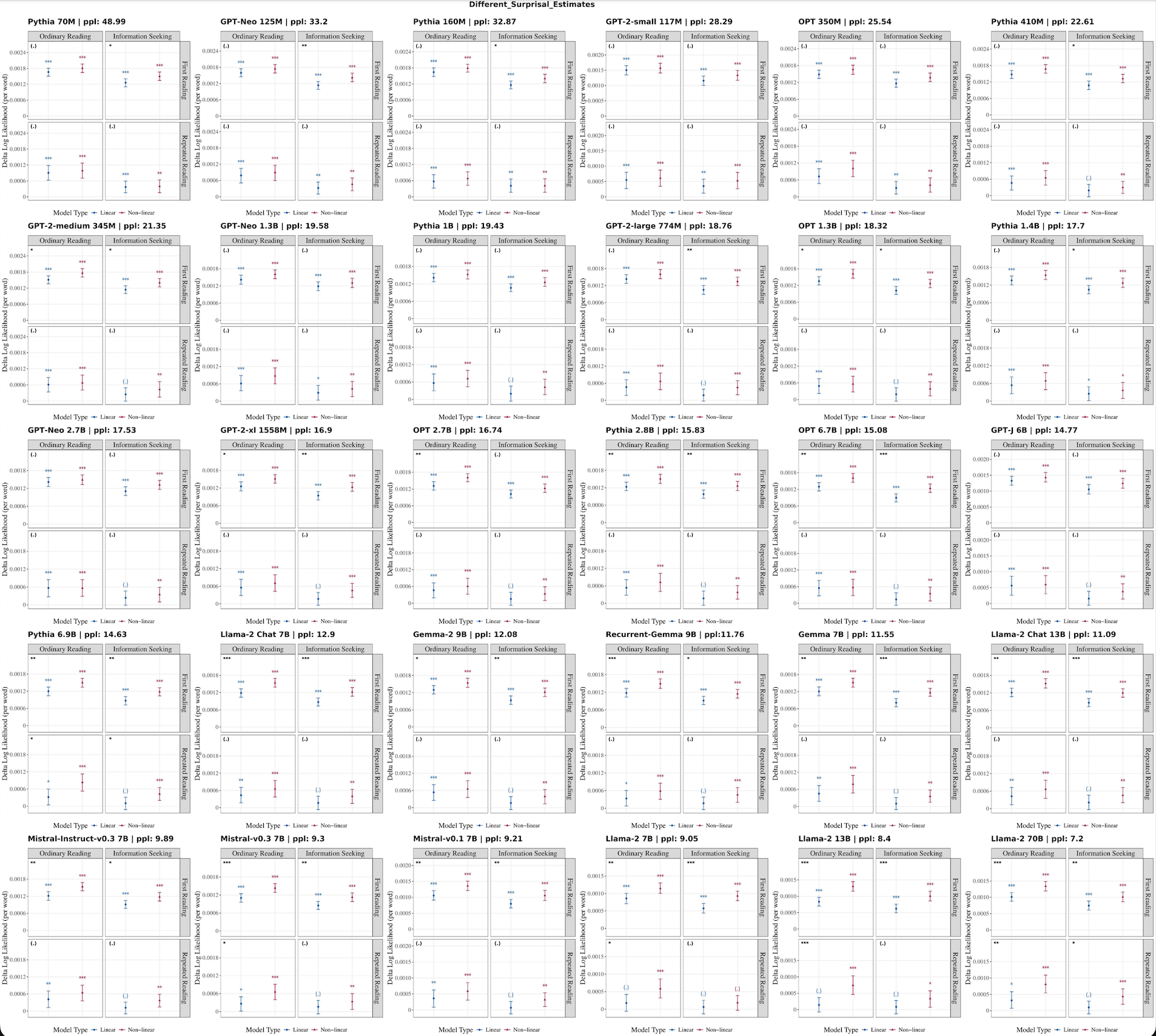}
        \caption{\textbf{Linearity Test: $\Delta LL$ means with 95\% confidence intervals on held-out data using 10-fold cross validation.} Above each confidence interval: the statistical significance of the $\Delta LL$ being different from zero, using a permutation test. Top left of each plot: statistical significance of a permutation test for a difference between the $\Delta LL$ of the linear and non-linear models.}
        \label{fig:Different Surprisal Estimates grid dll 1}
    \end{subfigure}
    \caption{{\textbf(a) GAM fits and (b) $\Delta LL$. Results for linear and non-linear models for \emph{different language models.}} `***' $p \leq 0.001$, `**' $p \leq 0.01$. `*' $p \leq 0.05$, `(.)' $p > 0.05$.}
    \label{fig:Different Surprisal Estimates grids}
\end{figure*}

\begin{figure*}[ht!]
    \centering
    \emph{Pythia-70m | Different Reading Measures}
    \begin{subfigure}[t]{0.49\textwidth}
        \centering
        \includegraphics[width=1.00\linewidth]{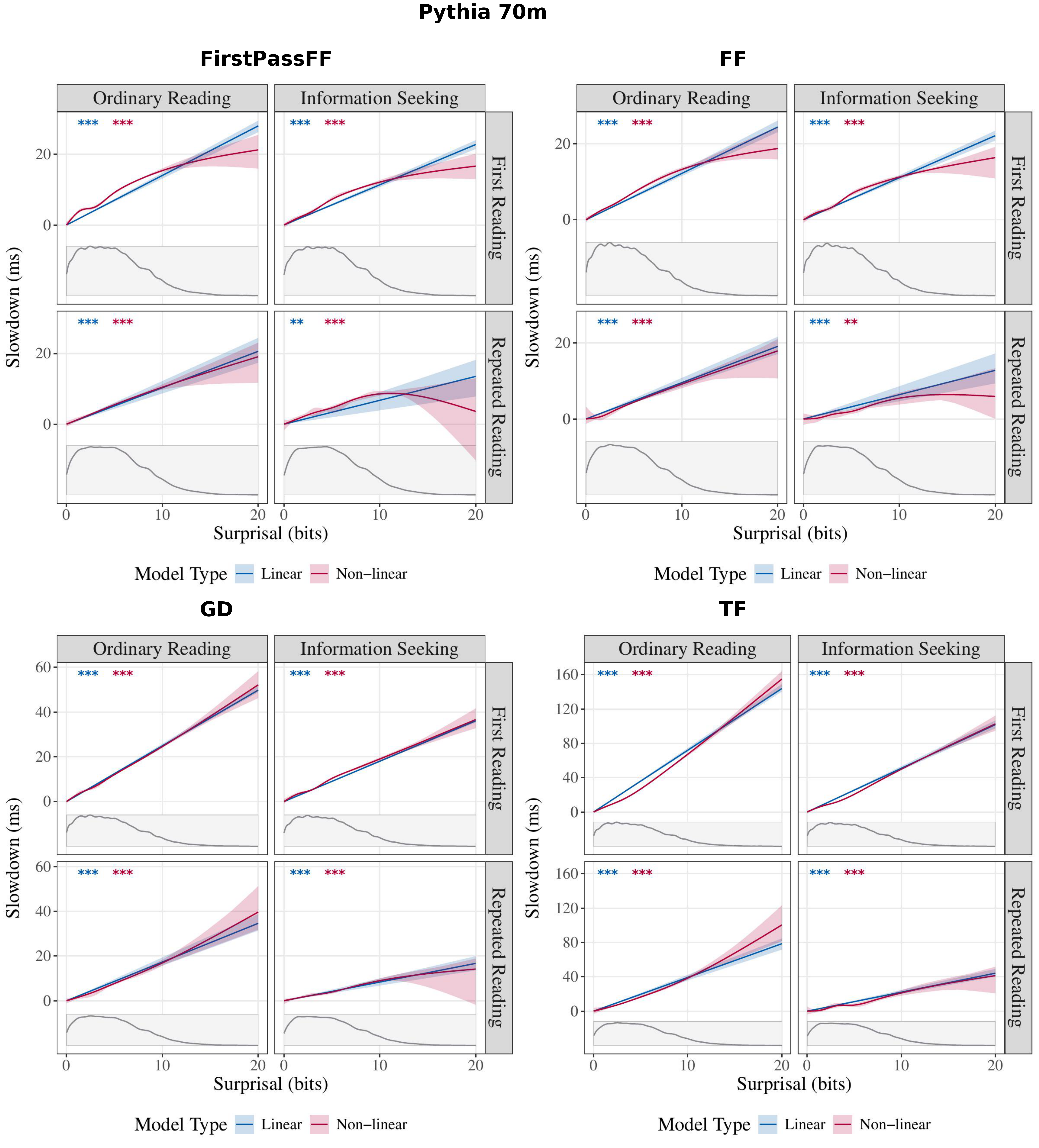}
        \caption{\textbf{GAM fits for the relation between surprisal and reading times, with bootstrapped 95\% confidence intervals.} Top left of each plot, the significance of the \texttt{s} and linear terms of the current word's surprisal. At the bottom of each plot: a density plot of surprisal values.}
        \label{fig:pythia-70m_grid_GAM}
    \end{subfigure}%
    ~ 
    \begin{subfigure}[t]{0.49\textwidth}
        \centering
        \includegraphics[width=1.00\linewidth]{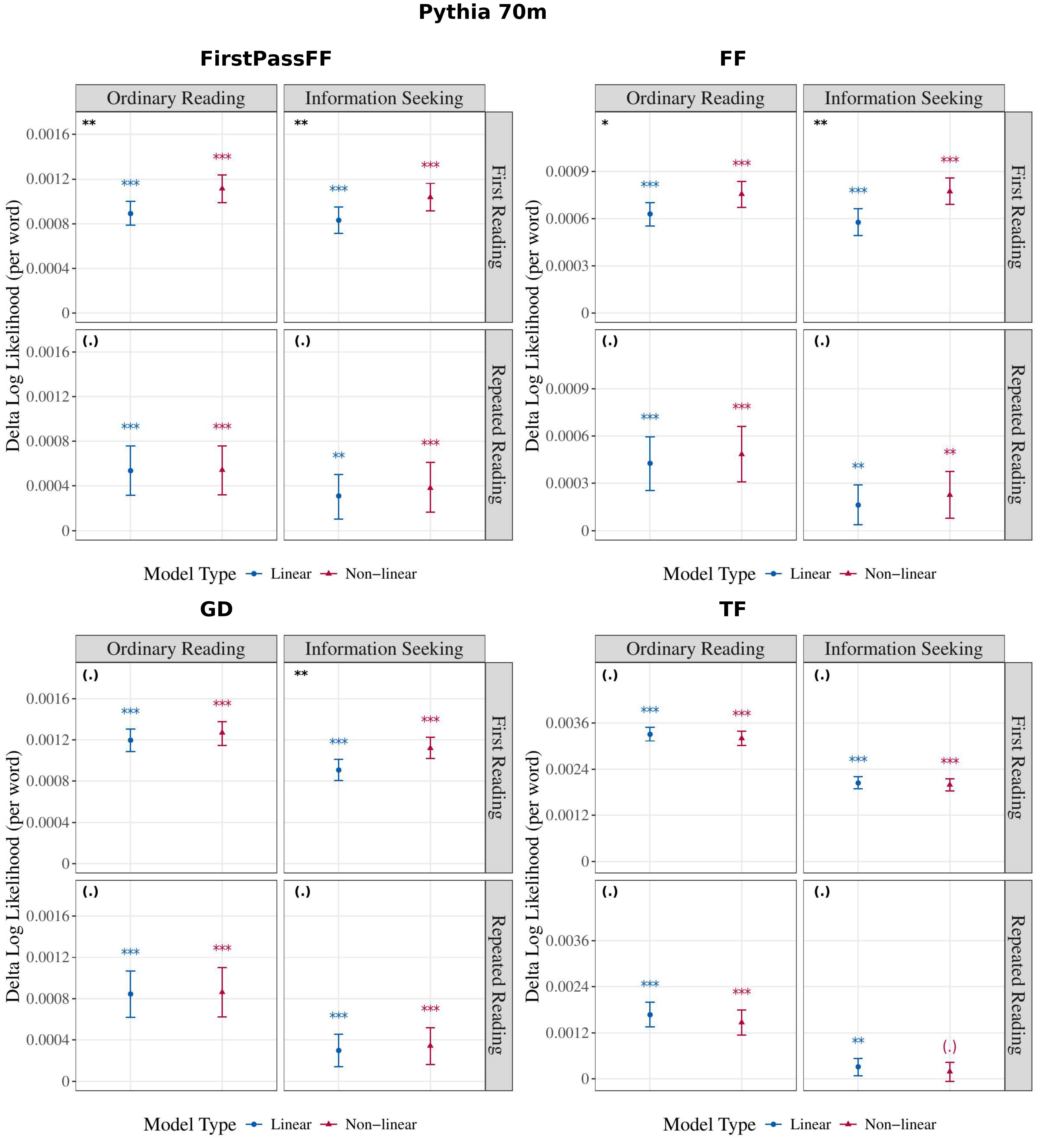}
        \caption{\textbf{Linearity Test: $\Delta LL$ means with 95\% confidence intervals on held-out data using 10-fold cross validation.} Above each confidence interval: the statistical significance of a permutation test that checks if the $\Delta LL$ is different from zero. Top left of each plot: statistical significance of a permutation test for a difference between the $\Delta LL$ of the linear and non-linear models.}
        \label{fig:pythia-70m_grid_dll}
    \end{subfigure}
    \caption{{\textbf(a) GAM fits and (b) $\Delta LL$ for different reading times, using  Surprisal estimates from \emph{Pythia-70m}.} `***' $p \leq 0.001$, `**' $p \leq 0.01$. `*' $p \leq 0.05$, `(.)' $p > 0.05$.}
    \label{fig:pythia-70m grids}
\end{figure*}

\begin{figure*}[ht!]
    \centering
    \emph{Pythia-70m | Data Subsets in Information Seeking and Repeated Reading}
    \begin{subfigure}[t]{\columnwidth}
        \centering
        \includegraphics[width=0.49\columnwidth]{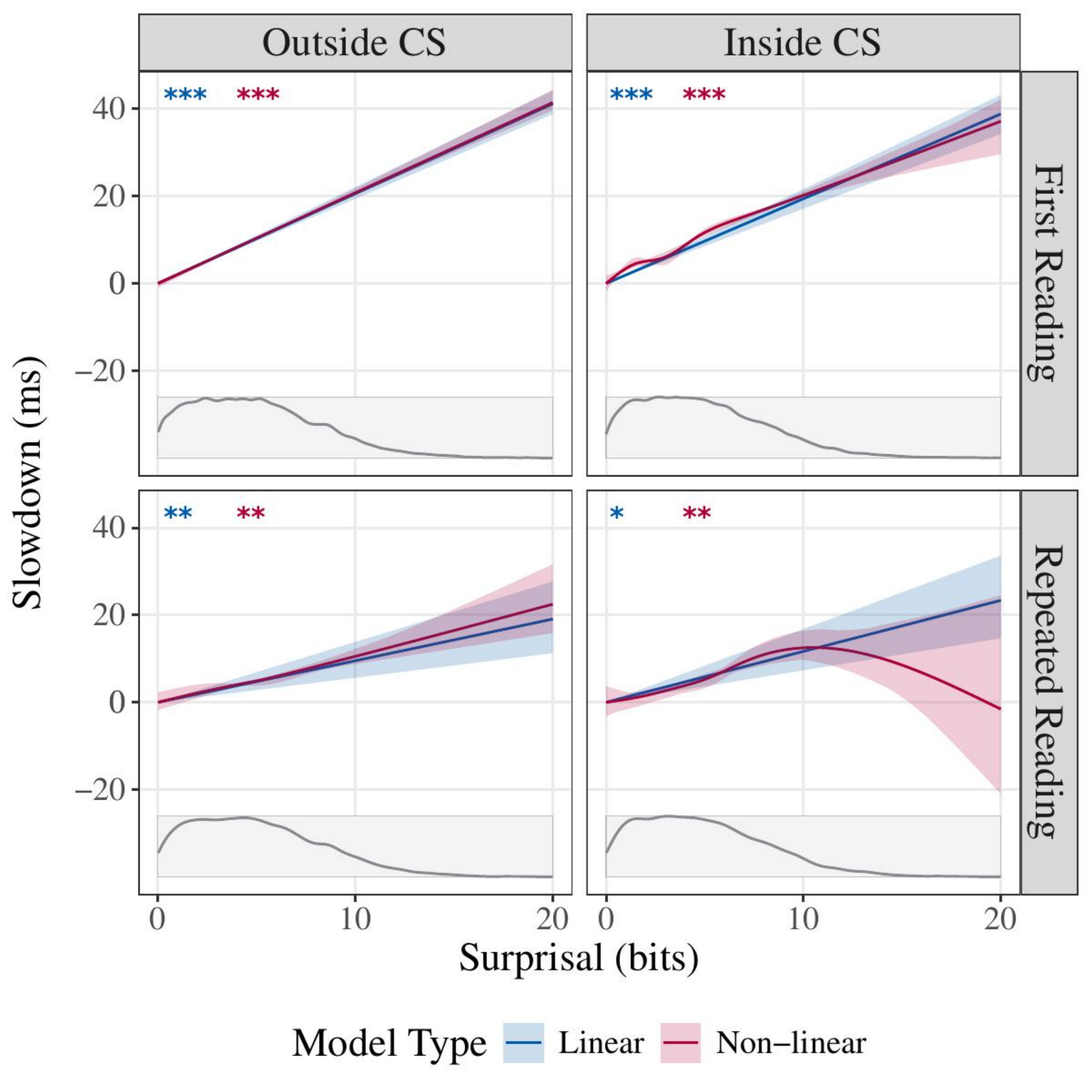}
        \hfill
        \includegraphics[width=0.49\columnwidth]{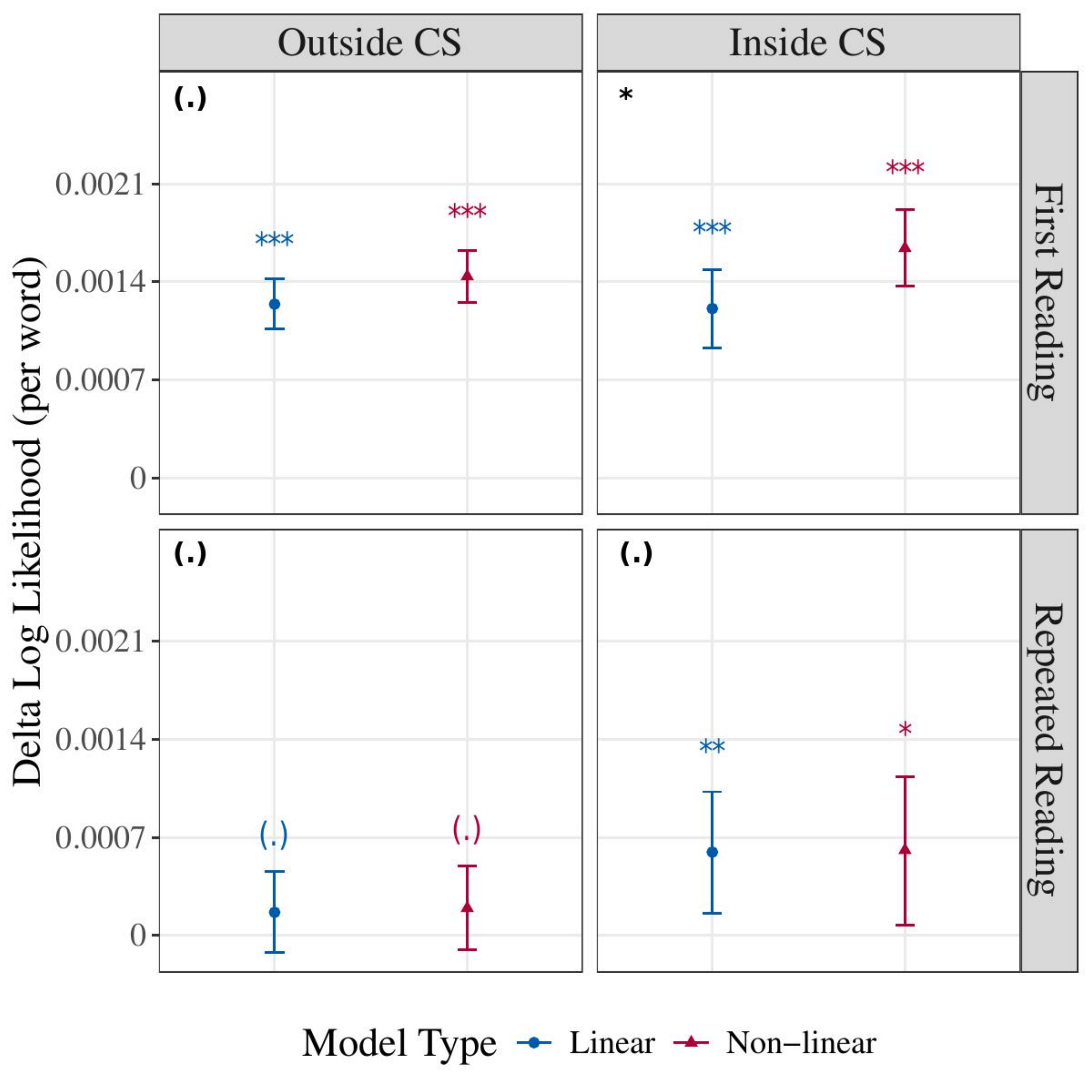}
        \caption{Information Seeking}
        \hspace{15mm}%
        \label{fig:FirstPassGD_GAM_within_outside_10_11}
    \end{subfigure}
    \begin{subfigure}[t]{\columnwidth}
        \centering        \includegraphics[width=0.49\columnwidth]{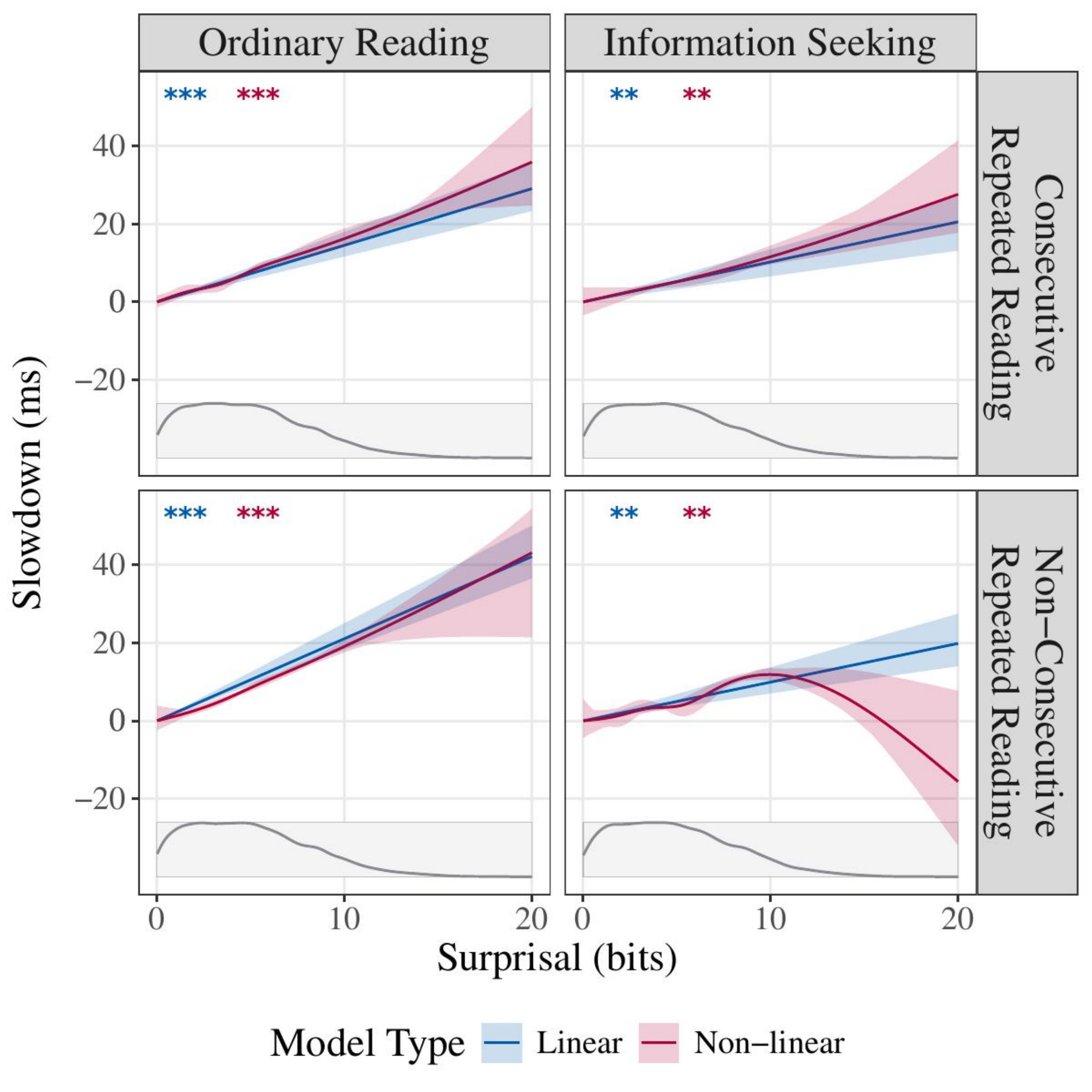}
        \includegraphics[width=0.49\columnwidth]{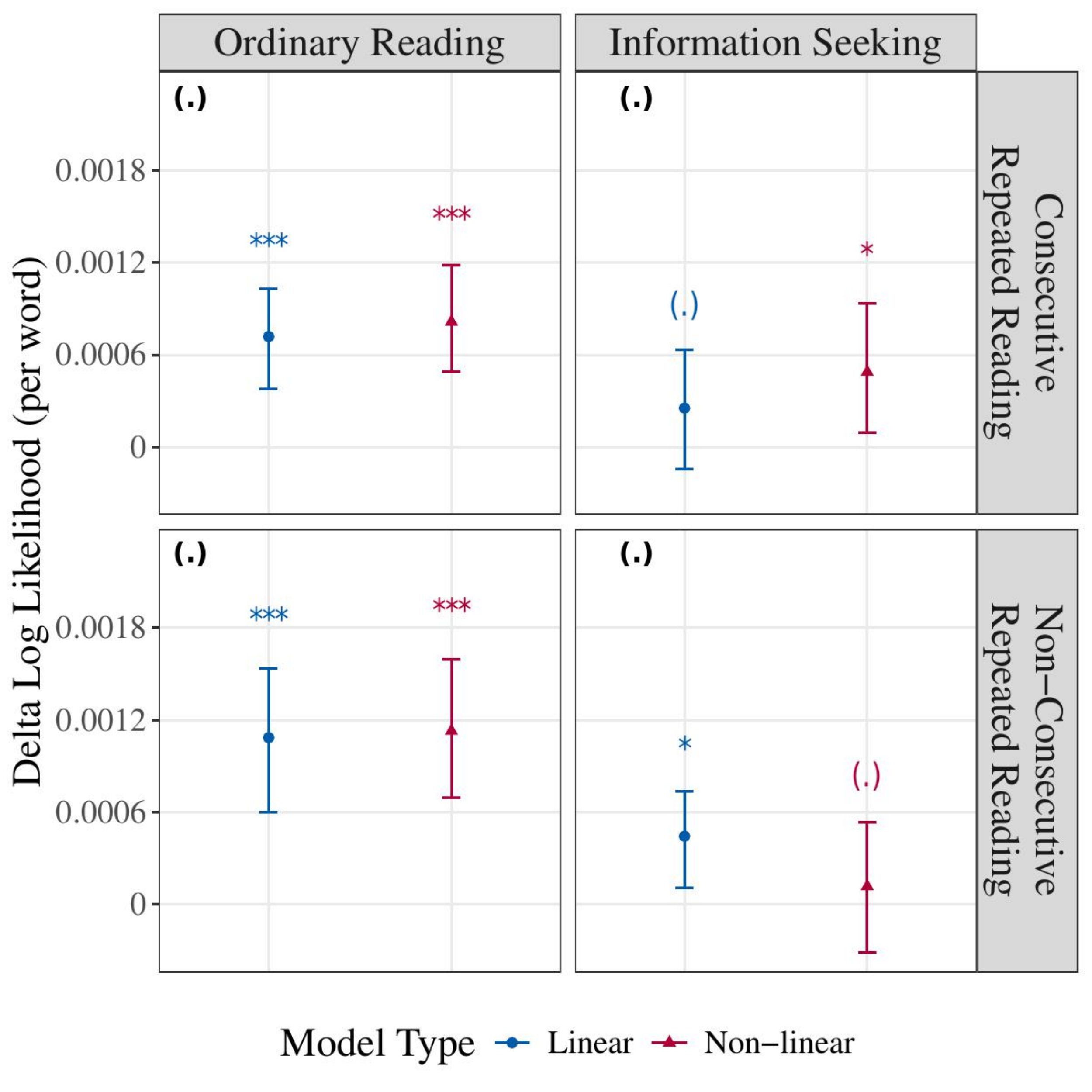}
        \caption{Repeated Reading}
        \label{fig:FirstPassGD_DLL_within_outside_10_11}

    \end{subfigure}
    \caption{GAM fits and $\Delta LL$ for first pass Gaze duration and Pythia-70m surprisals. (a) within versus outside the critical span (CS) in information seeking, and (b) consecutive (article 11) versus non-consecutive (article 12) repeated article reading. 
    }
    \label{fig:pythia70m-Hunting Reread splits}
\end{figure*}

\begin{figure*}[ht!]
    \centering
    \emph{GPT-2-small | Regime-specific Surprisal}
    \begin{subfigure}[t]{0.49\textwidth}
        \centering
        \includegraphics[width = 0.99\columnwidth]{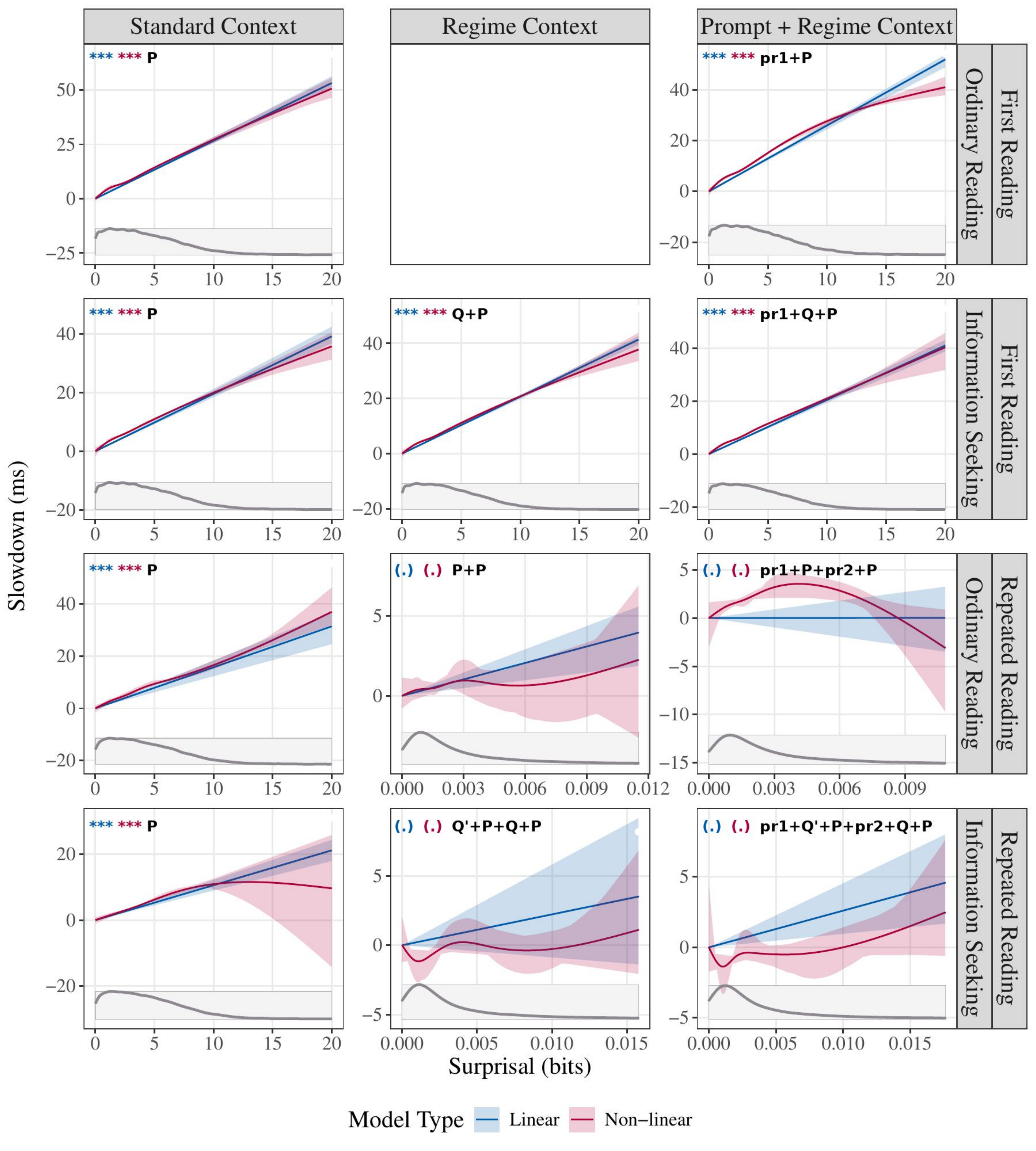}
        \caption{GAM fits for the relation between surprisal and reading times across context types. Slowdown effects in \textit{ms} as a function of  surprisal, with bootstrapped 95\% confidence intervals. Top left of each plot, the significance of the \texttt{s} and linear terms of the current word's surprisal. At the bottom of each plot: a density plot of surprisal values.}
    \label{fig:GPT2-small GAM_task_context}
    \end{subfigure}%
    ~ 
    \begin{subfigure}[t]{0.49\textwidth}
        \centering
        \includegraphics[width = 0.99\columnwidth]{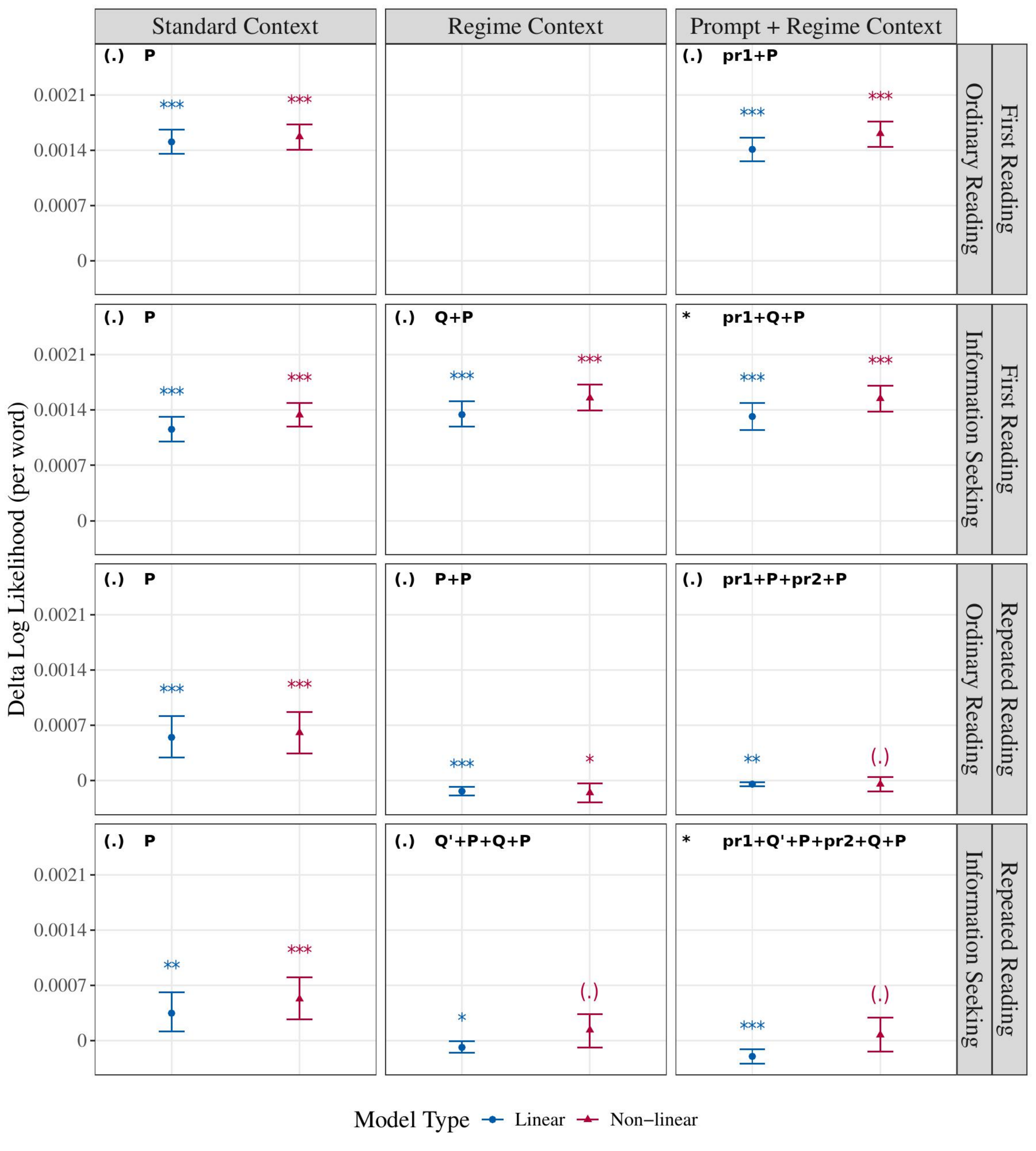}
        \caption{$\Delta LL$ means with 95\% confidence intervals on held-out data using 10-fold cross validation. Above each confidence interval: the statistical significance of a permutation test that checks if the $\Delta LL$ is different from zero. Top left of each plot: statistical significance of a permutation test for a difference between the $\Delta LL$ of the linear and non-linear models.}
        \label{fig:GPT2-small dll_context}
    \end{subfigure}
    \caption{\textbf{GPT-2-small} comparison of GAM fits and $\Delta LL$ for first pass Gaze Duration with surprisal estimates from different context types. `***' $p \leq 0.001$, `**' $p \leq 0.01$. `*' $p \leq 0.05$, `(.)' $p > 0.05$.}
\label{fig:GPT2-small context}
\end{figure*}

\begin{figure*}[ht!]
    \centering
    \emph{Llama 70b | Regime-specific Surprisal}
    \begin{subfigure}[t]{0.49\textwidth}
        \centering
        \includegraphics[width = 0.99\columnwidth]{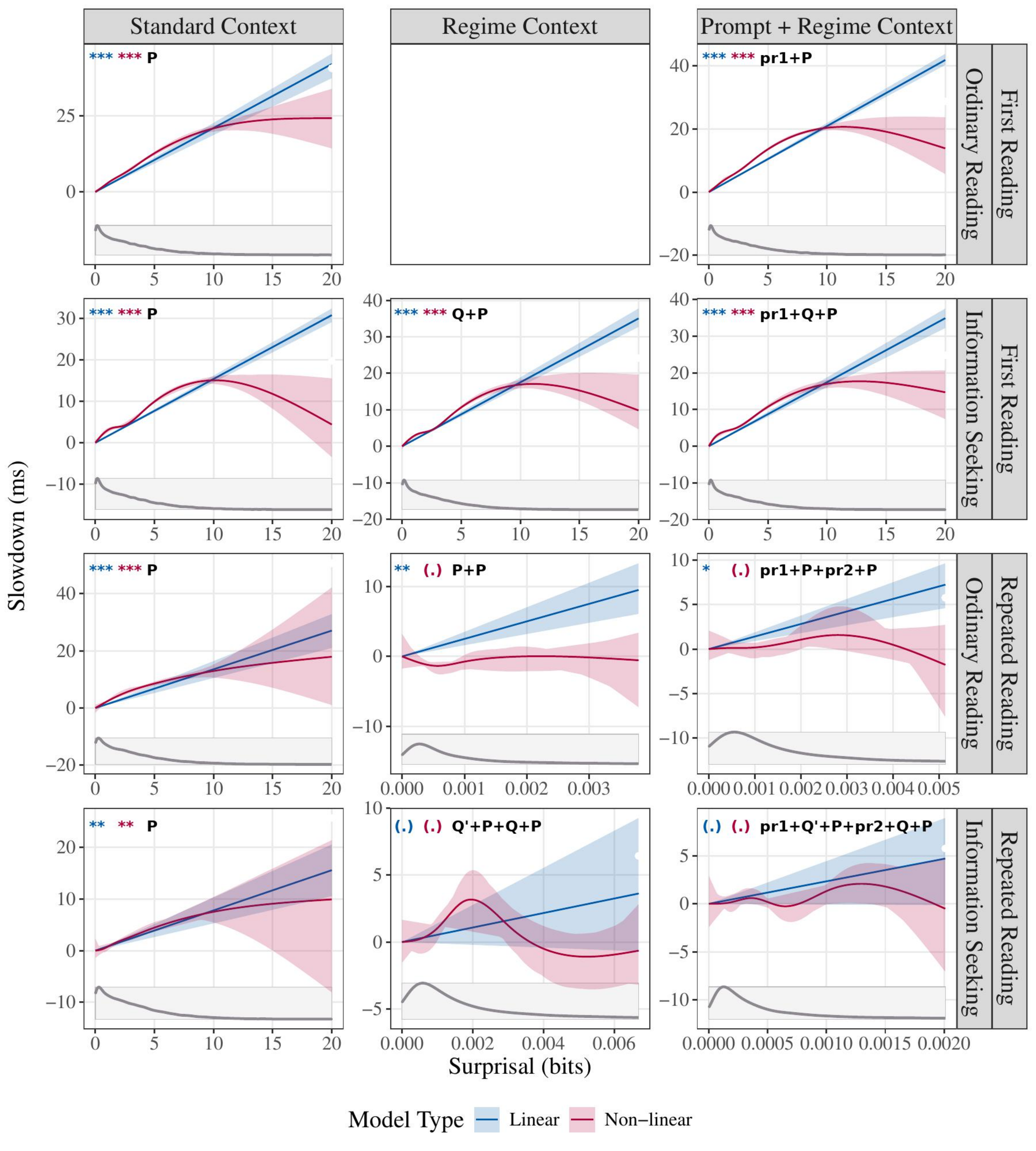}
        \caption{GAM fits for the relation between surprisal and reading times across context types. Slowdown effects in \textit{ms} for first pass Gaze Duration as a function of surprisal, with bootstrapped 95\% confidence intervals. Top left of each plot, the significance of the \texttt{s} and linear terms of the current word's surprisal. At the bottom of each plot: a density plot of surprisal values.}
    \label{fig:Llama 70b GAM_task_context}
    \end{subfigure}%
    ~ 
    \begin{subfigure}[t]{0.49\textwidth}
        \centering
        \includegraphics[width = 0.99\columnwidth]{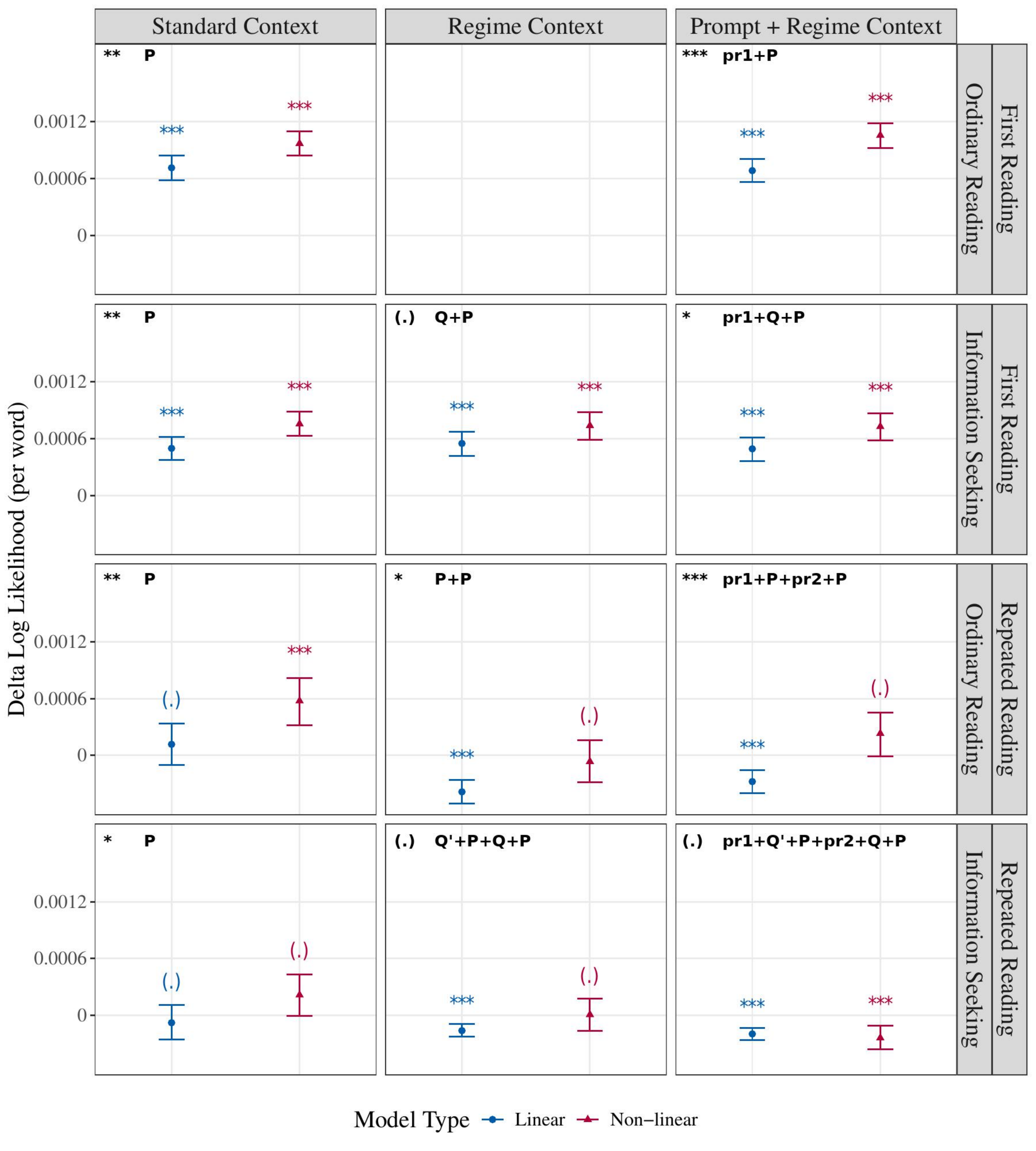}
        \caption{$\Delta LL$ means with 95\% confidence intervals on held-out data using 10-fold cross validation. Above each confidence interval: the statistical significance of a permutation test that checks if the $\Delta LL$ is different from zero. Top left of each plot: statistical significance of a permutation test for a difference between the $\Delta LL$ of the linear and non-linear models.}
        \label{fig:Llama 70b dll_context}
    \end{subfigure}
    \caption{\textbf{Llama 70b} comparison of GAM fits and $\Delta LL$ for first pass Gaze Duration with surprisal estimates from different context types. `***' $p \leq 0.001$, `**' $p \leq 0.01$. `*' $p \leq 0.05$, `(.)' $p > 0.05$. }
\label{fig:Llama 70b context}
\end{figure*}

\begin{figure*}[ht!]
    \centering
    \emph{Mistral Instruct v0.3 7b | Regime-specific Surprisal}
    \begin{subfigure}[t]{0.49\textwidth}
        \centering
        \includegraphics[width = 0.99\columnwidth]{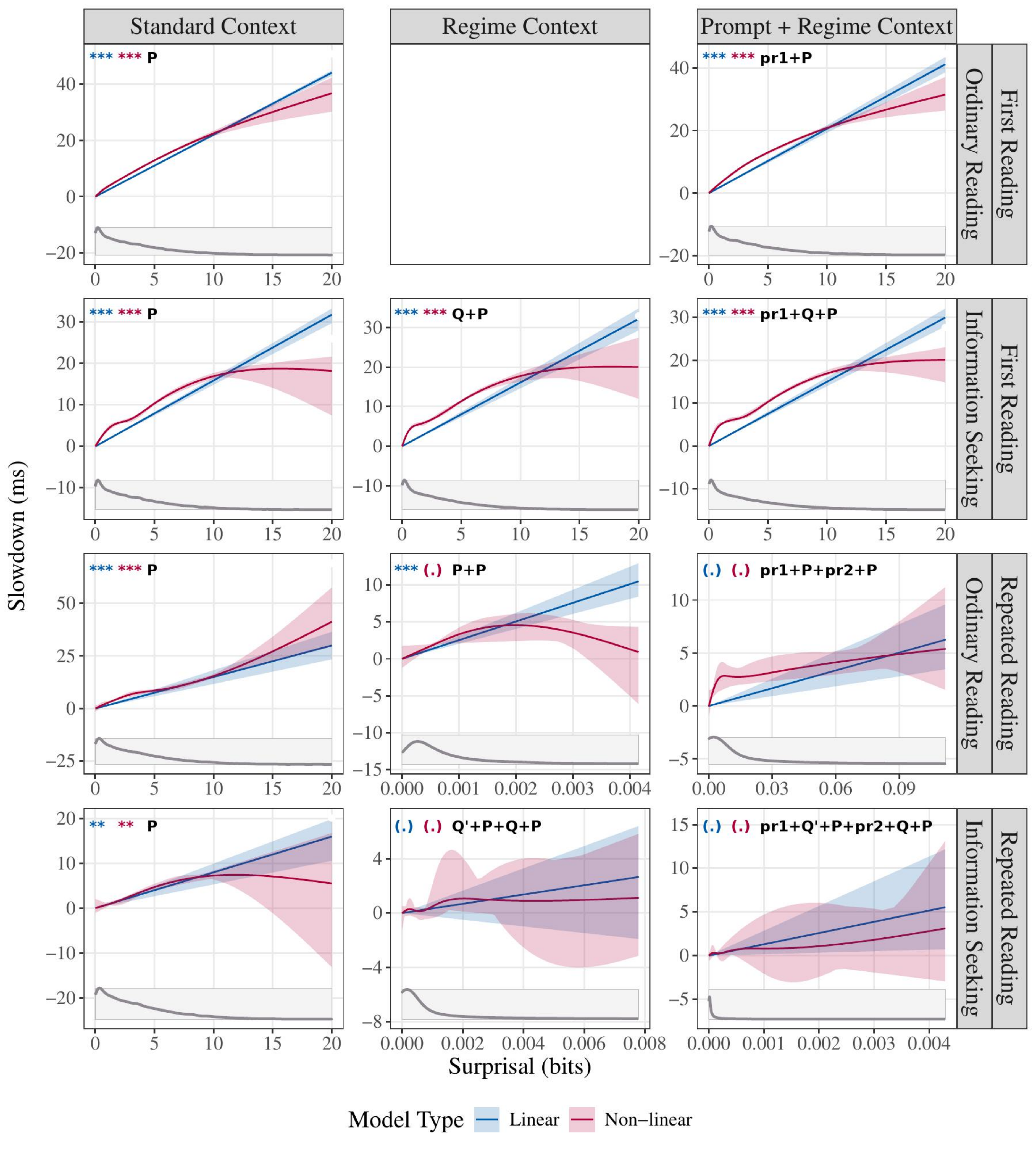}
        \caption{GAM fits for the relation between surprisal and reading times across context types. Slowdown effects in \textit{ms} for first pass Gaze Duration as a function of surprisal, with bootstrapped 95\% confidence intervals. Top left of each plot, the significance of the \texttt{s} and linear terms of the current word's surprisal. At the bottom of each plot: a density plot of surprisal values.}
    \label{fig:Mistral Instruct v0.3 7B GAM_task_context}
    \end{subfigure}%
    ~ 
    \begin{subfigure}[t]{0.49\textwidth}
        \centering
        \includegraphics[width = 0.99\columnwidth]{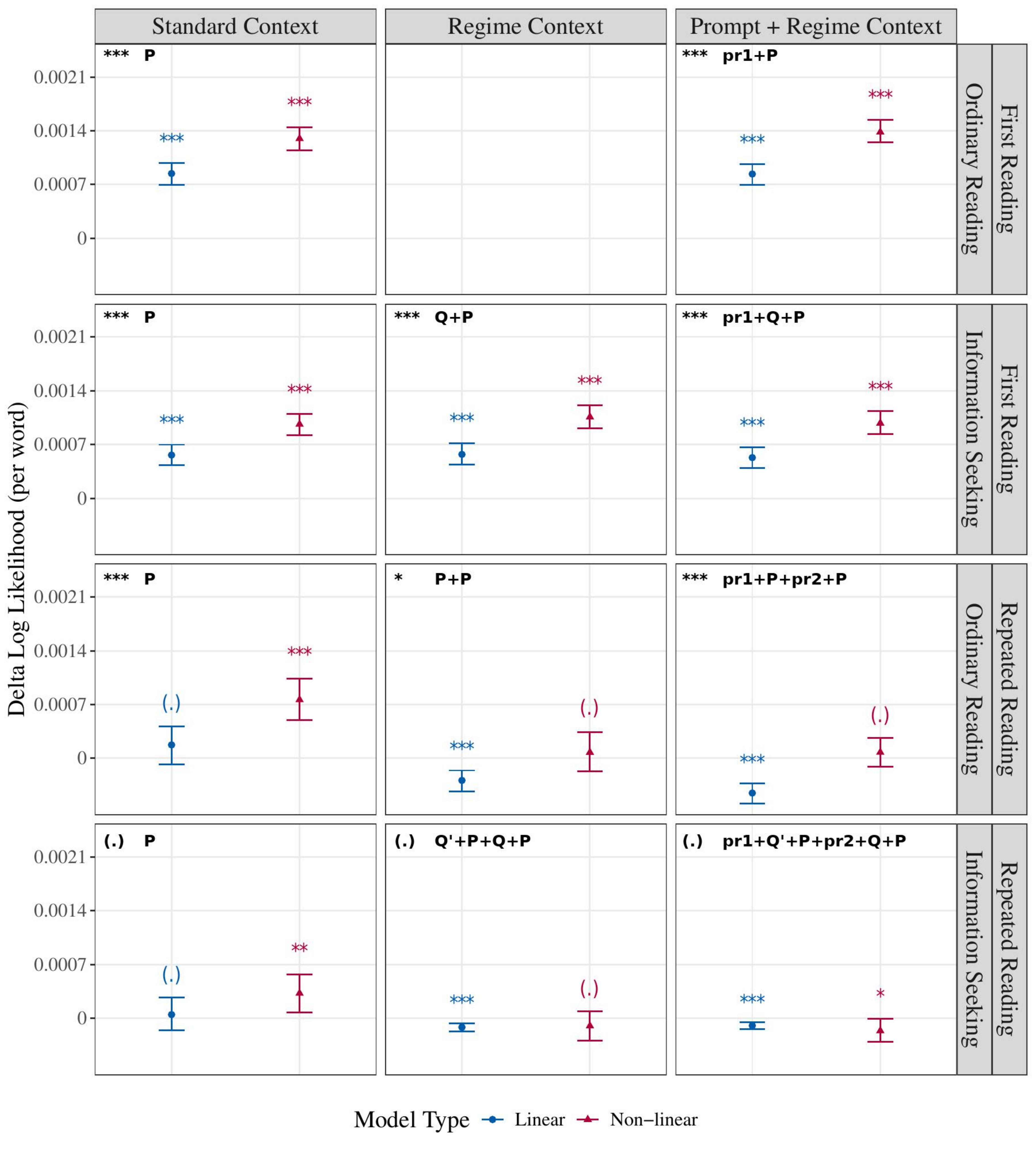}
        \caption{$\Delta LL$ means with 95\% confidence intervals on held-out data using 10-fold cross validation. Above each confidence interval: the statistical significance of a permutation test that checks if the $\Delta LL$ is different from zero. Top left of each plot: statistical significance of a permutation test for a difference between the $\Delta LL$ of the linear and non-linear models.}
        \label{fig:Mistral Instruct v0.3 7B dll_context}
    \end{subfigure}
    \caption{\textbf{Mistral Instruct v0.3 7b} comparison of GAM fits and $\Delta LL$ for first pass Gaze Duration with surprisal estimates from different context types. `***' $p \leq 0.001$, `**' $p \leq 0.01$. `*' $p \leq 0.05$, `(.)' $p > 0.05$. }
\label{fig:Mistral Instruct v0.3 7B context}
\end{figure*}

\begin{figure*}[ht!]
    \centering
    \emph{Pythia-70m | Article-level Regime-specific Surprisal}
    \begin{subfigure}[t]{0.49\textwidth}
        \centering
        \includegraphics[width = 0.99\columnwidth]{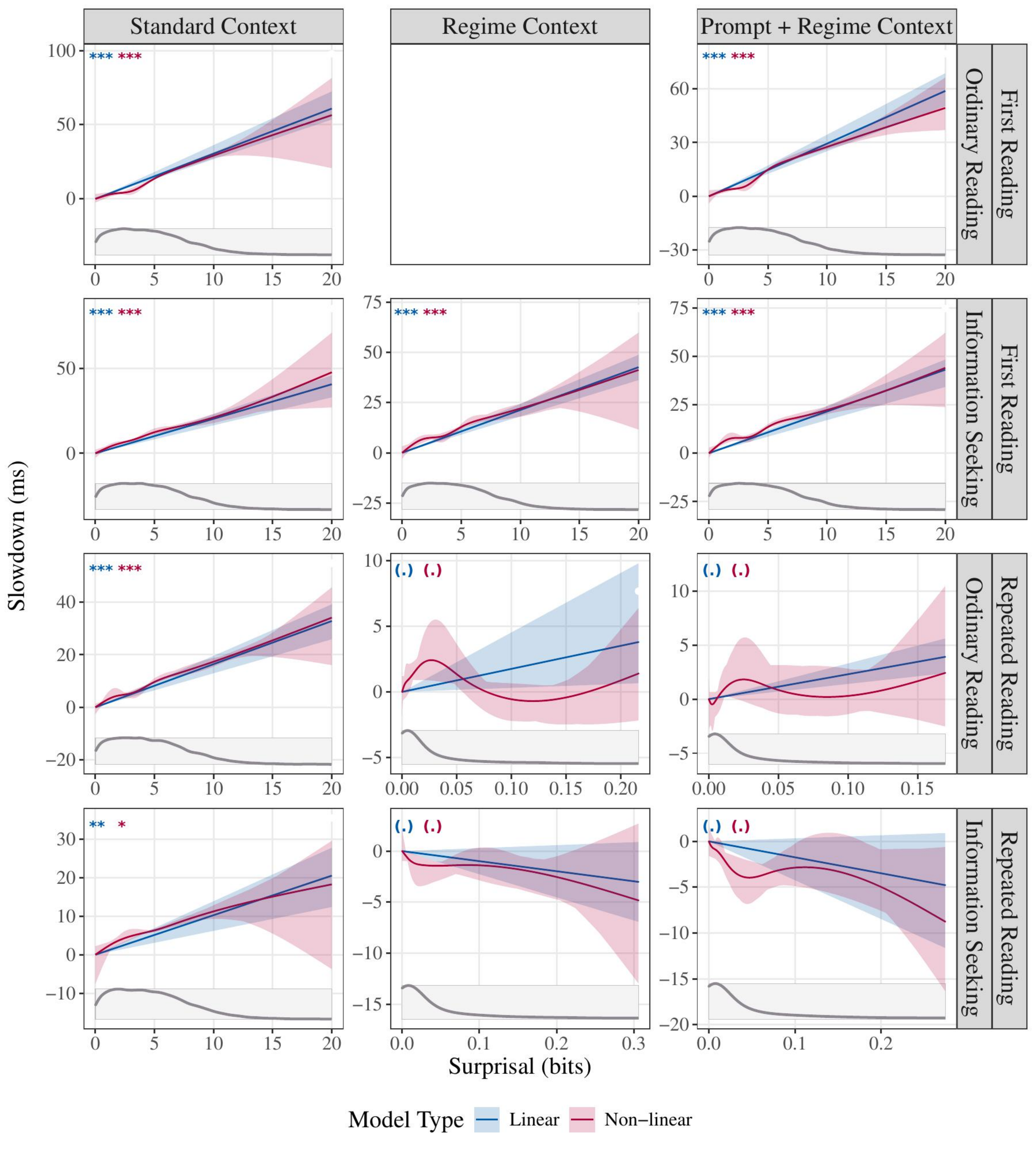}
        \caption{GAM fits for the relation between surprisal and reading times across context types. Slowdown effects in \textit{ms} for first pass Gaze Duration as a function of surprisal, with bootstrapped 95\% confidence intervals. Top left of each plot, the significance of the \texttt{s} and linear terms of the current word's surprisal. At the bottom of each plot: a density plot of surprisal values.}
    \label{fig:Pythia70m Article context gam}
    \end{subfigure}%
    ~ 
    \begin{subfigure}[t]{0.49\textwidth}
        \centering
        \includegraphics[width = 0.99\columnwidth]{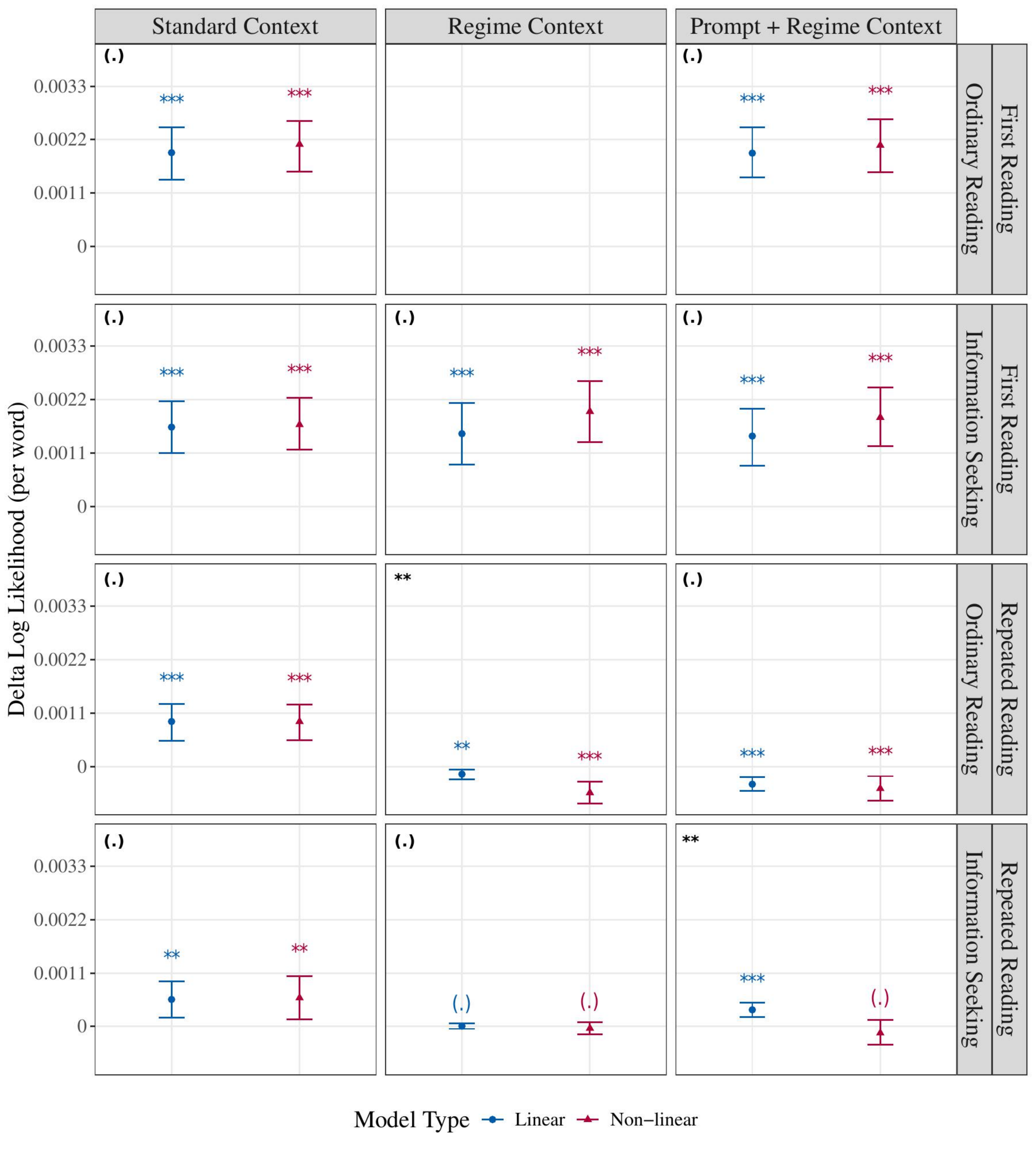}
        \caption{$\Delta LL$ means with 95\% confidence intervals on held-out data using 10-fold cross validation. Above each confidence interval: the statistical significance of a permutation test that checks if the $\Delta LL$ is different from zero. Top left of each plot: statistical significance of a permutation test for a difference between the $\Delta LL$ of the linear and non-linear models.}
        \label{fig:Pythia70m Article context dll}
    \end{subfigure}
    \caption{\textbf{Pythia-70m} comparison of GAM fits and $\Delta LL$ for first pass Gaze Duration with surprisal estimates from different \emph{article-level} context types. Regime-specific surprisal estimates in the Repeated Reading regime are based on the prior text \emph{including the intervening material} between the first and second readings of the current paragraph (as described in Table 1 in the Appendix). `***' $p \leq 0.001$, `**' $p \leq 0.01$. `*' $p \leq 0.05$, `(.)' $p > 0.05$. }
\label{fig:Pythia70m Article context}
\end{figure*}

\end{document}